\documentclass[runningheads]{llncs}
\usepackage{eccv}

\usepackage{eccvabbrv}

\usepackage{graphicx}
\usepackage{booktabs}

\usepackage[accsupp]{axessibility}  

\usepackage{hyperref}

\usepackage{orcidlink}
\usepackage{tikzducks}
\usepackage{graphicx}
\usepackage{amsmath}
\usepackage{amssymb}
\usepackage{booktabs}

\usepackage{color}
\usepackage{amssymb}
\usepackage{lipsum}
\usepackage{algorithm}
\usepackage{algpseudocode}
\usepackage{xcolor}
\usepackage{tabularx}
\usepackage{multirow}
\usepackage{enumitem}
\usepackage{bbm}
\usepackage{wrapfig}
\usepackage{setspace}
\usepackage{footnote}

\usepackage{subcaption}
\usepackage{colortbl} 
\usepackage{pifont}  
\usepackage{cite}  
\usepackage{mathtools}  
\usepackage[font=small]{caption}  
\usepackage{blindtext}  
\usepackage{stmaryrd} 
\usepackage[accsupp]{axessibility}  
\usepackage{afterpage}
\usepackage{ulem} 

\newcommand{\expnum}[2]{{#1}\mathrm{e}{-#2}}

\newcommand{\Fig}[1]{Fig.~\ref{fig:#1}}

\newcommand{\Tbl}[1]{Table~\ref{tab:#1}}

\def\ie{\textit{i.e.}}
\def\eg{\textit{e.g.}}

\definecolor{brown}{rgb}{0.85, 0.15, 0.15}
\definecolor{purp}{rgb}{0.95, 0.16, 0.65}
\definecolor{purpc}{rgb}{0.95, 0.36, 0.65}
\definecolor{orange}{rgb}{0.9, 0.45, 0.0}
\definecolor{blue}{rgb}{0.0, 0.5, 1.0}
\definecolor{green}{rgb}{0, 0.8, 0}
\definecolor{lgreen}{rgb}{0.6, 0.8, 0}
\definecolor{red}{rgb}{0.8, 0, 0}
\definecolor{redd}{rgb}{0.9, 0, 0}
\definecolor{yellow}{rgb}{0.75, 0.56, 0}
\definecolor{darkblue}{rgb}{0.2, 0.2, 0.8}
\definecolor{brinkpink}{rgb}{0.98, 0.38, 0.5}
\definecolor{cadmiumred}{rgb}{0.89, 0.0, 0.13}
\definecolor{ceruleanblue}{rgb}{0.16, 0.32, 0.75}
\definecolor{dandelion}{rgb}{0.94, 0.88, 0.19}
\definecolor{bostonuniversityred}{rgb}{0.8, 0.0, 0.0}
\definecolor{brown(web)}{rgb}{0.65, 0.16, 0.16}
\definecolor{cornellred}{rgb}{0.7, 0.11, 0.11}
\definecolor{greend}{rgb}{0.0, 0.35, 0.0}

\newcommand{\redd}[1]{{\color{redd}{#1}}}
\newcommand{\greend}[1]{{\color{greend}{#1}}}
\newcommand{\yellowd}[1]{{\color{yellow}{#1}}}

\def\etal{\textit{et al.}}

\definecolor{grey}{rgb}{0.9, 0.9, 0.9}

\newcommand{\ra}[1]{\renewcommand{\arraystretch}{#1}}

\newcommand{\hyperfootnote}[1][]{\def\ArgI\hyperfootnoteRelay}
\newcommand\hyperfootnoteRelay[2][]{\href{#1#2}{\ArgI}\footnote{\href{#1#2}{#2}}}

\begin{document}

\title{FREST: Feature RESToration for Semantic Segmentation under Multiple Adverse Conditions}
\titlerunning{FREST: Feature RESToration for Segmentation under Multiple Conditions}

\author{Sohyun Lee\inst{1}\orcidlink{0000-0003-4224-2019} \and
Namyup Kim\inst{2}\orcidlink{0000-0002-5503-2857} \and
Sungyeon Kim\inst{2}\orcidlink{0000-0002-6919-4822} \and
Suha Kwak\inst{1,2}\orcidlink{0000-0002-4567-9091}}

\authorrunning{S. Lee \etal}
\institute{Graduate School of Artificial Intelligence, POSTECH, Korea \and
Department of Computer Science and Engineering, POSTECH, Korea
{\tt\small \url{https://sohyun-l.github.io/frest}}
}

\maketitle
\begin{abstract}
Robust semantic segmentation under adverse conditions is crucial in real-world applications.
To address this challenging task in practical scenarios where labeled normal condition images are not accessible in training, we propose FREST, a novel feature restoration framework for source-free domain adaptation (SFDA) of semantic segmentation to adverse conditions.
FREST alternates two steps: (1) learning the condition embedding space that only separates the condition information from the features and (2) restoring features of adverse condition images on the learned condition embedding space.
By alternating these two steps, FREST gradually restores features where the effect of adverse conditions is reduced.
FREST achieved a state of the art on two public benchmarks (\ie, ACDC and RobotCar) for SFDA to adverse conditions.
Moreover, it shows superior generalization ability on unseen datasets.
\end{abstract}

\section{Introduction}\label{sec:intro}

The advent of deep neural networks has brought significant advancement of semantic segmentation~\cite{chen2018cascaded,sun2019deep,toshev2014deeppose,xiao2018simple,zhang2019fast}.
Although most existing models for semantic segmentation demonstrate outstanding performance under normal conditions, they often fail under real-world adverse conditions like fog, rain, snow, and nighttime that significantly degrade the quality of input images~\cite{Sakaridis_2018_ECCV,dai2020curriculum,Sakaridis_2019_ICCV,Zendel_2018_ECCV,Sakaridis_2018_IJCV,son2020urie,choi2021robustnet,lee2022fifo,lee2023human}. 
This lack of robustness limits the applicability of semantic segmentation, especially to high-stakes tasks like autonomous driving.

\setlength{\floatsep}{10pt plus 0pt minus 5pt}
\setlength{\textfloatsep}{10pt plus 0pt minus 5pt}
\begin{figure}
    \centering
    \includegraphics[width=\linewidth]{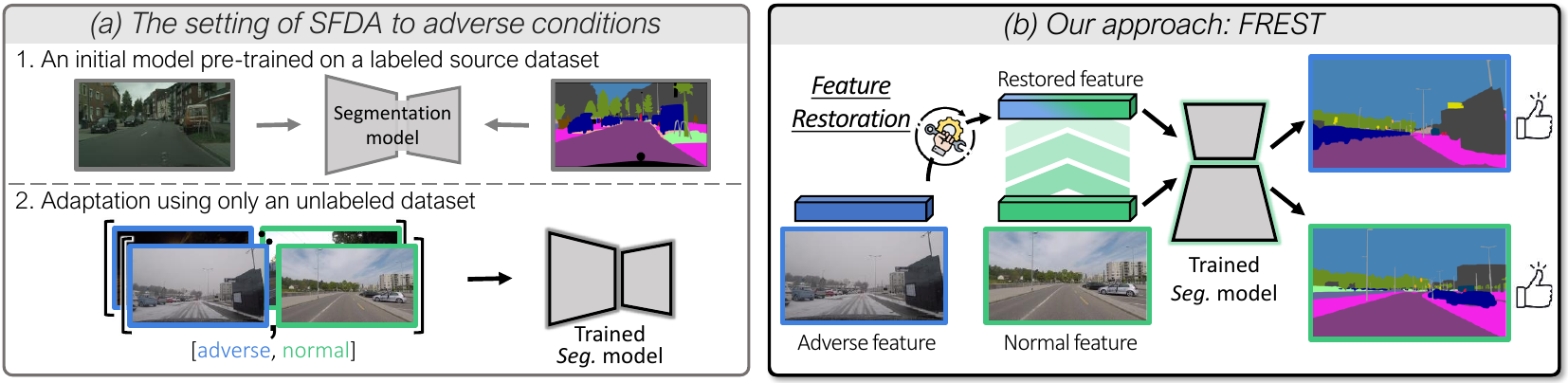}
    \caption{
    (a) The setting of SFDA to adverse conditions. 
    A segmentation model, initially pre-trained on a labeled source dataset, is adapted to adverse conditions using pairs of unlabeled adverse and normal images.
    (b) Following the SFDA setting, FREST restores features of adverse condition images to simulate the normal condition. 
    }
    \label{fig:teaser}
\end{figure}

An obstacle in enhancing the robustness of semantic segmentation models is the difficulty of collecting labeled data for every possible adverse condition. This issue has steered the computer vision community towards unsupervised domain adaptation (UDA)~\cite{hoffman2016fcns, zhang2017curriculum, tsai2018learning, tsai2019domain, vu2019advent,zou2019confidence, pan2020unsupervised, kim2020learning,hoyer2022daformer, hoyer2022hrda, bruggemann2023refign, hoyer2023mic}. UDA is the task of training a model using labeled data from a \textit{source} domain (\ie~clear weather) and unlabeled data from a \textit{target} domain (\ie~adverse weather conditions) while bridging the gap between these domains.
This approach mitigates the need for labeling target domain data while improving performance in that domain.

More recent research has explored source-free domain adaptation (SFDA), a more practical form of UDA where access to the source domain data is not allowed due to privacy leaks or the prohibitively large scale of the data.~\cite{wang2021tent, fleuret2021uncertainty,liu2021source,guo2022simt,zhao2023towards,bruggemann2023contrastive}.
In SFDA, a model is first pre-trained with labeled data from the source domain and then fine-tuned using only unlabeled data from the target domain afterwards.
In particular, SFDA to adverse conditions has been studied on a specialized setup~\cite{bruggemann2023contrastive}, in which target domain images are taken under various adverse conditions and each target image is paired with a reference image taken in a similar geolocation but under the normal condition, \ie, clear weather.
This setting differs from the conventional SFDA as the reference images from the normal condition are available, but following the previous work~\cite{bruggemann2023contrastive}, we refer to this setting as SFDA to adverse conditions. 
In this setting, a pair of target and reference images are matched by global navigation satellite system (GNSS), and thus are only roughly aligned due to the variations between them in terms of camera pose and shooting time.
Also, both target and reference images are unlabeled, and the adverse condition type of each target image is unknown. 
This setup is illustrated in Fig.~\ref{fig:teaser}(a).

The prior work for SFDA to adverse conditions enhances the robustness of semantic segmentation models by learning condition-invariant features~\cite{bruggemann2023contrastive}. 
To this end, it encourages features extracted from each pair of adverse (target) and normal (reference) images to be close.
However, regarding that normal images could resemble the source domain, updating features of normal images to be close to those of adverse images causes the model pre-trained on the source domain to forget the rich knowledge of the domain, resulting in poor representations for both normal and adverse images in the end.
Also, the feature matching relies heavily on the assumption that the alignment between a pair of adverse and normal images is sufficiently accurate, which typically does not hold, unfortunately.

In this work, we introduce a novel framework for \textit{Feature RESToration for multiple adverse conditions}, called \textbf{FREST}, which is illustrated in \Fig{teaser}(b).
FREST overcomes the aforementioned limitations by restoring features of adverse condition images so that they simulate the normal condition of reference images in feature spaces.
This notion of feature restoration is embodied by extracting and leveraging \textit{condition-specific information}, which ideally depends only on the condition of image and is not affected by its semantic content.
The use of condition-specific information enables our method to restore features while considering only the condition of input.
Hence, FREST mitigates the catastrophic forgetting of the source domain knowledge through the feature restoration, and is less affected by the content mismatch between a pair of adverse and normal images by utilizing condition-specific information.

FREST operates in two steps as follows.
First, it learns a new embedding space that represents only condition-specific attributes of images.
To be specific, in this embedding space, images taken under similar conditions are closely aligned while those under different conditions are separate.
We consider the embedding vector of an image in this space as its condition-specific information.
In the second stage, FREST learns restored features for adverse images, tailored for semantic segmentation, by optimizing the model for both segmentation and feature restoration.
In detail, the segmentation objective leverages pseudo segmentation labels as supervision, while the objective for feature restoration enforces condition-specific information of adverse images approximates that of their corresponding normal images so that \textit{the model learns to represent images of any conditions as if they are taken under the normal condition}.
FREST alternates between these two stages so that condition-specific information is adapted to reflect the update of the segmentation network in the first stage, which improves the restored features of the model in the second stage consequently. 
It is empirically demonstrated in Sec.~\ref{sec:empirical} that FREST effectively learns restored features of adverse images.

FREST was evaluated on the standard benchmarks for SFDA to adverse conditions based on the Cityscapes~\cite{cityscapes}, ACDC~\cite{acdc}, and RobotCar~\cite{larsson2019cross,maddern20171} datasets,
\ie, Cityscapes $\to$ ACDC and Cityscapes $\to$ RobotCar settings.
FREST achieved a new state of the art in both two settings.
Moreover, its superiority in terms of robustness and generalization capability was demonstrated on unseen datasets by applying our model for Cityscapes$\to$ACDC to the ACG~\cite{bruggemann2023contrastive} and Cityscapes-lindau40~\cite{cityscapes} datasets.

\section{Related Work}
\label{sec:relate}

\noindent\textbf{Robustness.}
Robust recognition has been actively studied due to its relation with crucial safety-critical applications~\cite{Sakaridis_2018_ECCV,dai2020curriculum,Sakaridis_2019_ICCV,Zendel_2018_ECCV,Sakaridis_2018_IJCV,son2020urie,choi2021robustnet,lee2022fifo,lee2023human}.
In this context, various UDA methods~\cite{hoyer2022daformer, hoyer2022hrda, bruggemann2023refign, hoyer2023mic} have been proposed to improve robustness across multiple adverse conditions.
In particular, Br\"uggemann~\etal~\cite{bruggemann2023refign} suggest a method of spatial alignment between target and reference images and adaptive label correction guided by the warping results.
Br\"uggemann~\etal~\cite{bruggemann2023contrastive} introduce a method for SFDA under multiple adverse conditions through contrastive learning for condition-invariant learning.
In contrast to the previous work, FREST restores features from multiple adverse conditions to those of the normal condition, effectively learning robust features for adverse conditions.

\vspace{0.5mm}
\noindent\textbf{Unsupervised Domain Adaptation.}
UDA has been widely studied for semantic segmentation with the introduction of synthetic datasets~\cite{gta5,synthia}, which provide automatically generated pixel-level labels. 
UDA allows the use of both labeled synthetic and unlabeled real data for training.
Existing UDA methods are mainly categorized into distribution alignment~\cite{hoffman2016fcns, zhang2017curriculum, tsai2018learning, tsai2019domain, vu2019advent} and self-training~\cite{zou2018unsupervised, li2019bidirectional, zou2019confidence, pan2020unsupervised, kim2020learning}.
Revealing the potential of transformers~\cite{vaswani2017attention, dosovitskiy2020image, xie2021segformer} for semantic segmentation, recent studies have conducted transformer-based approaches~\cite{hoyer2022daformer, hoyer2022hrda, bruggemann2023refign, hoyer2023mic}.
Despite significant improvements, UDA for semantic segmentation faces the practical limitation of requiring access to labeled source data for adaptation. 
To address this, source-free domain adaptation is introduced, using only unlabeled target data to adapt a source-trained model.

\vspace{0.5mm}
\noindent\textbf{Source-free Domain Adaptation.}
SFDA is introduced to adapt a source-trained model to the target domain without accessing source data.
The early approaches suggest test-time objectives at the output space of an unlabeled target domain dataset by entropy minimization~\cite{wang2021tent, fleuret2021uncertainty}, data-free knowledge distillation~\cite{liu2021source}, and contrastive learning~\cite{huang2021model}.
Recently, Guo~\etal~\cite{guo2022simt} introduce a plug-and-play method via a noise transition matrix for loss correction on noisy pseudo-labeled target data.
Zhao~\etal~\cite{zhao2023towards} suggest enhancing the stability and adaptability of self-training through a dynamic teacher update mechanism and a resampling strategy based on training consistency.
Motivated by this line of research, Bruggemann~\etal~\cite{bruggemann2023contrastive} propose a practical setup, where target images are taken under multiple adverse conditions and associated with reference images of the normal condition.
Following this setup, FREST learns to restore adverse features using normal features for robust semantic segmentation.

\section{Configuration of Target Domain Data} 
\label{sec:dataset_config}

Following the problem setting of the previous work~\cite{bruggemann2023contrastive}, we assume that target domain data comprise images taken under various adverse conditions, and that each target image is paired with a reference image captured under the normal condition.
Both target and reference images are unlabeled, and it is unknown under what condition each target image was taken.
Moreover, a target adverse condition image $I_\textrm{adv}$ and its associated reference normal condition image $I_\textrm{norm}$ are matched by GNSS so that they are taken in similar geolocations.
Note however that a pair of GNSS-matched images will only be roughly aligned due to the variations between them in terms of camera pose and shooting time. 
To mitigate this issue, following prior work~\cite{bruggemann2023refign,bruggemann2023contrastive}, we warp ${I}_\textrm{norm}$ onto ${I}_\textrm{adv}$ using the UAWarpC dense matching network~\cite{bruggemann2023refign} pre-trained on the MegaDepth dataset~\cite{li2018megadepth}.
More details for the warping process can be found in~\cite{bruggemann2023refign}.
While the warping alleviates the misalignment issue to some extent, it still leaves nontrivial discrepancy in content due to imperfect warping and dynamic objects.

\section{Proposed Method}
\label{sec:method}

\setlength{\floatsep}{10pt plus 0pt minus 5pt}
\setlength{\textfloatsep}{10pt plus 0pt minus 5pt}
\begin{figure*}[t]
    \centering
    \includegraphics[width=1\linewidth]{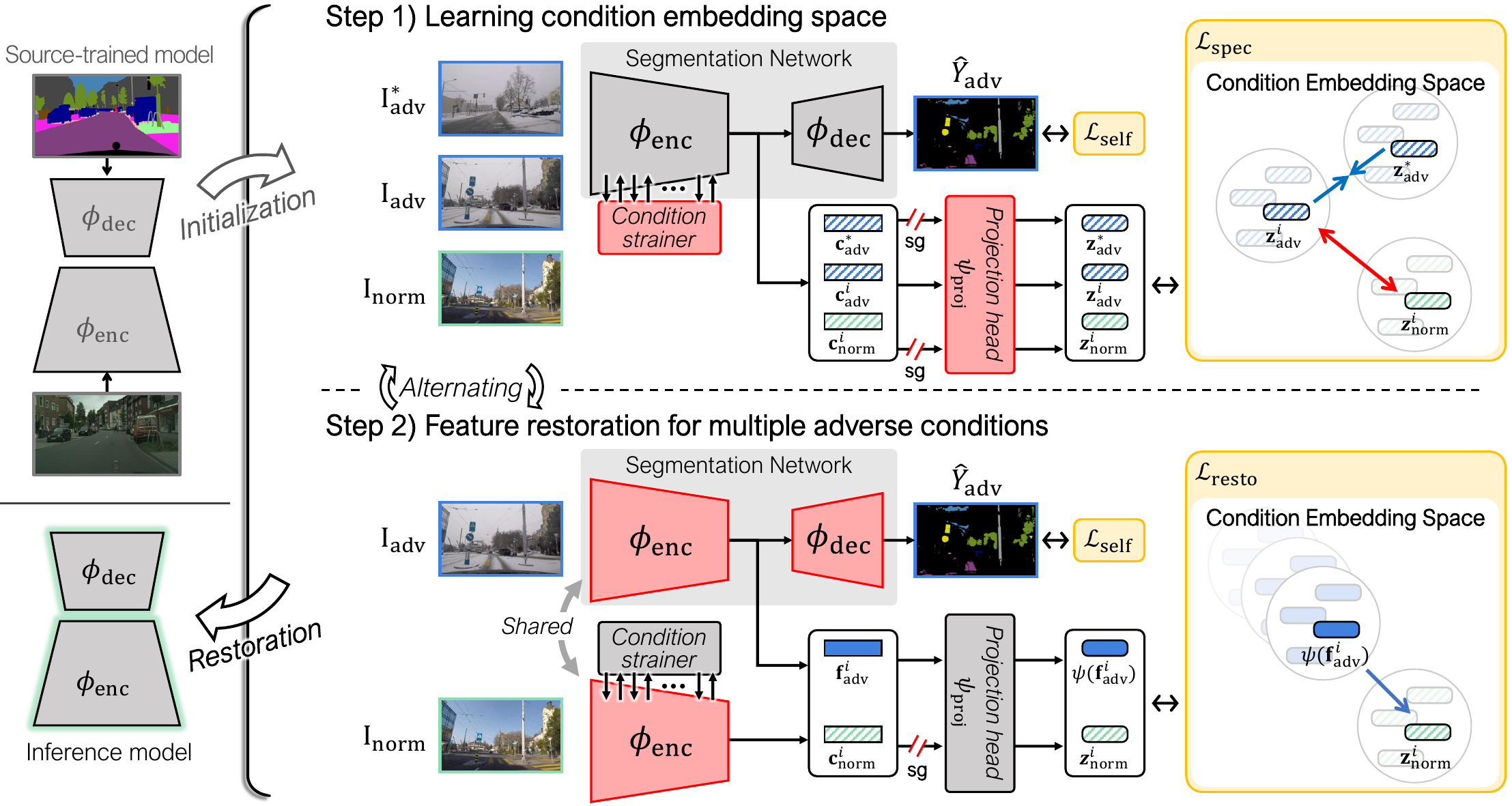}
    \caption{
    The overall architecture and training strategy.
    The segmentation network is pre-trained using a labeled source dataset.
    For each iteration, the condition strainer and segmentation network are trained alternatingly.
    The frozen modules are shown in gray, the trainable modules are highlighted in red, and ``sg'' denotes the stop gradient.
    (\textit{Step 1}) The condition strainer and projection head are trained to learn the condition embedding space.
    (\textit{Step 2}) The segmentation network is trained to restore features from adverse to normal conditions on the condition embedding space.
    For evaluation, only the encoder $\phi_\textrm{enc}$ and decoder $\phi_\textrm{dec}$ of the segmentation network are utilized.
    }
\label{fig:pipeline}
\end{figure*}

The overall architecture and training strategy of our model are outlined in \Fig{pipeline}; we suppose that the segmentation network is pre-trained with a labeled source domain dataset following the standard protocol~\cite{bruggemann2023contrastive}.
FREST considers adverse conditions as detrimental, and aims to remove their effects in features of a target adverse image $I_\textrm{adv}$ by \textit{feature restoration}, which is the process of learning features of $I_\textrm{adv}$ that resemble those of the corresponding normal image $I_\textrm{norm}$, not in the content, but only in the effect of the condition.

However, using features of $I_\textrm{norm}$ and $I_\textrm{adv}$ directly for feature restoration is less than ideal, as it may lead to a distortion of the semantic content, due to the misalignment between $I_\textrm{norm}$ and $I_\textrm{adv}$ discussed in Sec.~\ref{sec:dataset_config}.
To address this issue, FREST extracts condition-specific information from each feature and utilizes it to guide the feature restoration process.
To realize this idea, FREST alternates the following two steps: (1) learning a condition embedding space, and (2) restoring features of $I_\textrm{adv}$ so that the features approximate those of $I_\textrm{norm}$ on the condition embedding space.

In the first step, FREST learns a condition embedding space to capture the condition-specific information of input.
In this space, images are separated and clustered based on whether they were taken under adverse conditions or the normal condition.
We consider the embedding vector of an image in this space as its condition-specific information, which is less affected by its semantic content.
For learning such an embedding space, we propose to attach a module named \textit{condition strainer}, denoted by $\psi_\textrm{strainer}$, to the frozen segmentation encoder $\phi_\textrm{enc}$ so that the module retains only condition-specific information that is distinct from the source domain information preserved by the frozen encoder.

\setlength{\floatsep}{10pt plus 0pt minus 5pt}
\setlength{\textfloatsep}{10pt plus 0pt minus 5pt}
\begin{wrapfigure}{br}{0.36\linewidth}
  \centering
  \begin{minipage}{1.0\linewidth}
        \centering
        \includegraphics[width=0.9\textwidth]{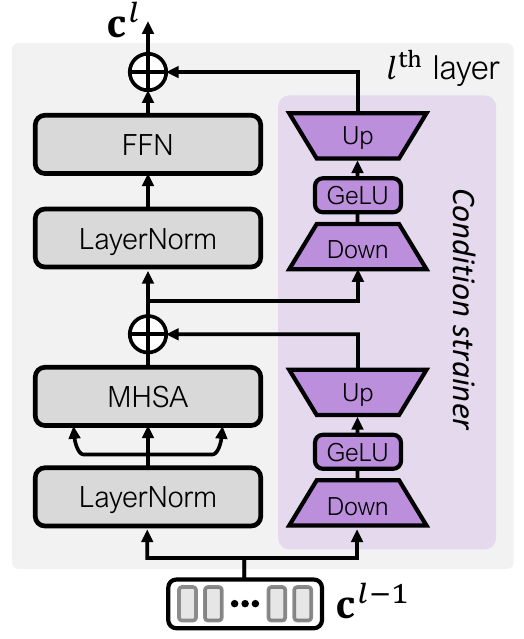}
    \caption{
    Detail of the encoder with the condition strainer.
    Condition strainers are connected to the original feed-forward layer (FFN) and multi-head self-attention layer (MHSA) through the residual connections.
    }
    \label{fig:detail}
  \end{minipage}
\end{wrapfigure}

The design of the condition strainer is inspired by parameter-efficient fine-tuning~\cite{houlsby2019parameter,adaptformer,pfeiffer2020adapterfusion}, which adds a small number of parameters for capturing task-specific information. 
This structure enables the condition strainer to capture condition-specific information effectively and efficiently.
The detailed architecture of the encoder with the condition strainer is presented in \Fig{detail}.
The condition strainer is trained to extract condition-specific information from each layer of the encoder while being separated from the encoder; it is separate from the encoder so that the encoder is not affected by its condition-specific information.
We call features extracted by $\phi_\textrm{enc}$ incorporating $\psi_\textrm{strainer}$ condition-infused features.
Specifically, the condition-infused features $\textbf{c}^{l}$ computed at the $l^\textrm{th}$ layer of the encoder is given by $\textbf{c}^l=\phi_{\textrm{enc}}^l(\textbf{c}^{l-1})+\psi_\textrm{strainer}^l(\textbf{c}^{l-1})$.
We indicate $\textbf{c}$ as the condition-infused feature produced by the last layer of the encoder.
Finally, $\textbf{c}$ is projected on the condition embedding space through a projection head $\psi_\textrm{proj}$.
Also, for the sake of brevity, we denote encoder features computed only by $\phi_{\textrm{enc}}$, disregarding the strainer $\psi_\textrm{strainer}$, by $\textbf{f}^l=\phi_{\textrm{enc}}^l(\textbf{f}^{l-1})$ where $\textbf{f}$ is the encoder feature obtained from the last layer.
The encoder features $\textbf{f}^l$ are targeted features for the feature restoration in the second step.

In the second step, we train the segmentation network while conducting the feature restoration with the frozen condition strainer and projection head.
FREST restores features of $I_\text{adv}$ from $\phi_{\textrm{enc}}$, denoted by $\textbf{f}_\text{adv}$, to resemble condition-infused features of $I_\text{norm}$, denoted by $\textbf{c}_\textrm{norm}$, where $\textbf{f}_\text{adv}$ and $\textbf{c}_\textrm{norm}$ are computed from the last layer of the encoder.
Specifically, $\textbf{f}_\textrm{adv}$ is encouraged to approximate $\textbf{c}_\textrm{norm}$ by a regression objective on the condition embedding space for considering only the condition information during feature restoration.

By alternating the two steps aforementioned, the segmentation network progressively learns restored features for multiple adverse conditions, and the condition strainer is adapted to reflect the update of the segmentation network and facilitates the next feature restoration consequently. The remainder of this section elaborates on the two steps of FREST.

\subsection{Learning Condition Embedding Space}\label{sec:spec_loss}
In the first step, FREST learns the condition embedding space that only represents condition-specific information of input.
To this end, we train the condition strainer and projection head with the frozen segmentation network.
Loss functions used in this step are described below.

\vspace{2mm}\noindent\textbf{Condition-specific Learning Loss.}
The goal of condition-specific learning is to learn the condition embedding space by extracting only condition-specific information from the condition-infused features that contain both content and conditions.
We implement this objective by contrastive learning, grouping features from the same condition closely together and distancing those from different conditions.
For contrastive learning, the anchor and positive are sampled with different semantics under the same condition, while the anchor and negative are chosen to be semantically similar but under different conditions.

To sample a pair of anchor and negative, we warp normal features to corresponding adverse features as described in Sec.~\ref{sec:dataset_config}.
Specifically, based on the confidence scores computed by the warping module~\cite{bruggemann2023refign}, we select patch embeddings surpassing a warping confidence threshold of 0.2 for computing the anchor and negative.
Given a pair of $I_\textrm{norm}$ and $I_\textrm{adv}$, we first sample $\textbf{c}_{\textrm{norm}}^{i}$ and ${\textbf{c}}_{\textrm{adv}}^{i}$ as condition-infused features for $i^\textrm{th}$ patch embedding, for $i \in \mathcal{W}$ where $\mathcal{W} = \{i~|~\mathrm{conf}(i)\geq0.2\}$ with $\mathrm{conf}(i)$ indicating the warping confidence score.
\setlength{\floatsep}{10pt plus 0pt minus 5pt}
\setlength{\textfloatsep}{10pt plus 0pt minus 5pt}
\begin{figure}[t!]
\centering
  \begin{minipage}{0.56\linewidth}
        \includegraphics[width=0.98\textwidth]{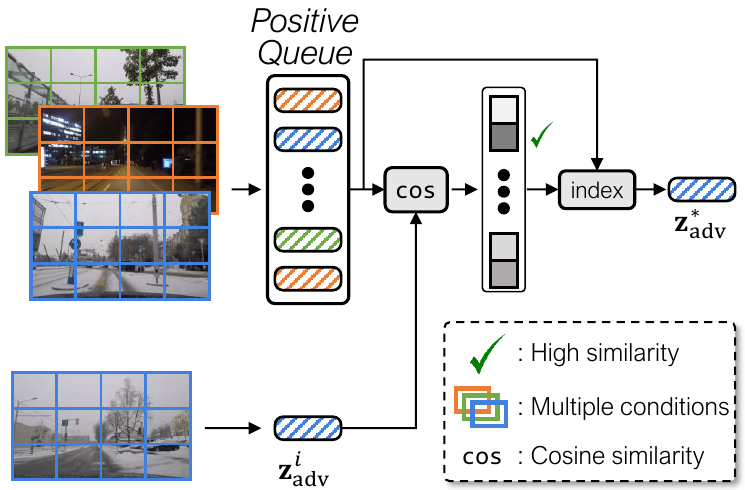}
    \caption{
    Detail of positive embedding sampling strategy in the condition-specific learning.
    }
    \label{fig:pos_select}
  \end{minipage}
  \quad 
  \begin{minipage}{0.34\linewidth}
        \centering
        \includegraphics[width=0.8\textwidth]{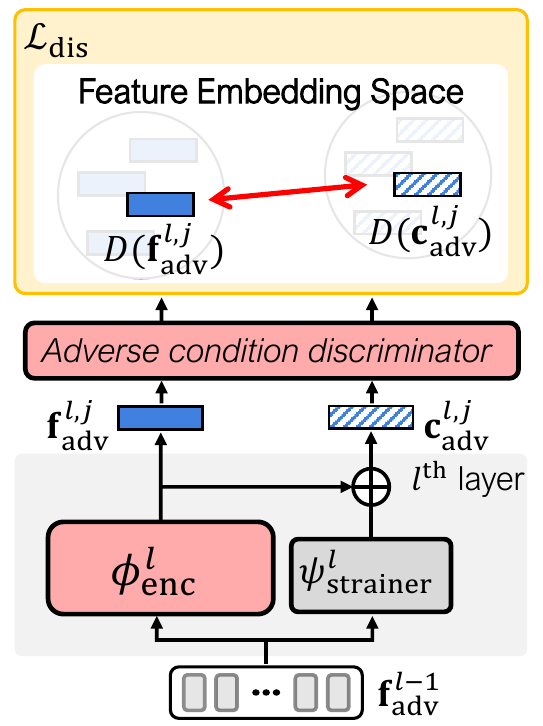}
    \caption{Detail of the adverse condition discriminator.
    }
    \label{fig:loss_dis}
  \end{minipage}
\end{figure}
Then, the anchor and negative samples are computed by projecting $\textbf{c}_{\textrm{adv}}^{i}$ and ${\textbf{c}}_{\textrm{norm}}^{i}$ on the condition embedding space;
$\textbf{z}_{\textrm{adv}}^{i}=\psi_\textrm{proj}(\textbf{c}_{\textrm{adv}}^{i})$ and ${\textbf{z}}_{\textrm{norm}}^{i}=\psi_\textrm{proj}({\textbf{c}}_{\textrm{norm}}^{i})$, respectively.
To obtain sufficient positive candidates, we stack condition embeddings of adverse images in a batch into a positive queue of length $Q$ for each iteration.
As shown in \Fig{pos_select}, we choose the representative positive ${\textbf{z}}_{\textrm{adv}}^{*}$ that is the most similar to the anchor $\textbf{z}_{\textrm{adv}}^{i}$ in the queue, assuming it has the same condition with the anchor; we empirically verify that ${\textbf{z}}_{\textrm{adv}}^{*}$ is an effective reference point for pushing a negative sample $\textbf{z}_{\textrm{norm}}^{i}$ in Sec.~\ref{sec:patch}.
For each patch embedding, the condition-specific loss aims to attract the anchor $\textbf{z}_{\textrm{adv}}^{i}$ towards the representative positive ${\textbf{z}}_{\textrm{adv}}^{*}$ and repel it from the negative ${\textbf{z}}_{\textrm{norm}}^{i}$.

For $\{\textbf{z}_{\textrm{adv}}^{i}, {\textbf{z}}_{\textrm{adv}}^{*}, \textbf{z}_{\textrm{norm}}^{i}\}$, the condition-specific loss for patch $i$ is given by:
\begin{align}
&\mathcal{L}_{\textrm{spec},i}= -\log\frac{\exp({\textbf{z}_{\textrm{adv}}^{i}}^\top{\textbf{z}}_{\textrm{adv}}^{*}/\tau)}{\exp({\textbf{z}_{\textrm{adv}}^{i}}^\top{\textbf{z}}_{\textrm{adv}}^{*}/\tau))+\exp({\textbf{z}_{\textrm{adv}}^{i}}^\top{\textbf{z}}_{\textrm{norm}}^{i}/\tau)},
\label{eq:cf_loss}
\end{align}
where $\tau$ serves as a temperature that scales the sensitivity.
The condition-specific loss is the average of the losses in \eqref{eq:cf_loss} applied to individual patches with warping confidence scores larger than the threshold, and formulated as
\begin{align}
\mathcal{L}_{\textrm{spec}} = \frac{1}{|\mathcal{W}|}\sum\limits_{i \in \mathcal{W}}\mathcal{L}_{\textrm{spec}, i}.
\label{eq:sum_cf_loss}
\end{align}

\vspace{2mm}\noindent\textbf{Total Loss for Step 1.} 
Following the previous work~\cite{bruggemann2023contrastive}, we employ a self-training loss $\mathcal{L}_{\textrm{self}}$, \ie, the pixel-wise cross-entropy loss with pseudo labels generated by class-balanced self-training (CBST)~\cite{zou2018unsupervised}. 
It is necessary for learning semantic information due to the absence of segmentation labels during training.
To preserve the semantic information of condition-infused features, we apply the self-training loss to the predictions computed from condition-infused features.
Then, the total training loss of Step 1 is represented as $\mathcal{L}_\textrm{step1}=\lambda_{\textrm{spec}}\mathcal{L}_{\textrm{spec}}+\mathcal{L}_{\textrm{self}}$.

\subsection{Learning Semantic Segmentation with Feature Restoration}
In the second step, FREST learns to restore features so that $\textbf{f}_\textrm{adv}$ from adverse conditions approximates $\textbf{c}_\textrm{norm}$ from the normal condition in the condition embedding space.
To this end, we train the segmentation network with the frozen condition strainer $\psi_\textrm{strainer}$.
In addition, for further alleviating adverse effects from $\textbf{f}_\textrm{adv}^l$, FREST ensures that $\textbf{f}_\textrm{adv}^l$ is discriminated from the adverse condition-infused feature $\textbf{c}_\textrm{adv}^l$ for the $l^\textrm{th}$ layer of the encoder.
The loss functions in this process are described below.

\vspace{2mm}\noindent\textbf{Feature Restoration Loss.}
Given paired ${I}_\textrm{adv}$ and ${I}_\textrm{norm}$, we compute $\textbf{f}_\textrm{adv}$ using $\phi_{\textrm{enc}}$ only and a condition-infused feature ${\textbf{c}}_{\textrm{norm}}$.
Note that we prevent gradient updates from ${\textbf{c}}_{\textrm{norm}}$ to ensure that the adverse condition one-sidedly follows the normal condition as ${\textbf{c}}_{\textrm{norm}}$ is the target for restoration of $\textbf{f}_\textrm{adv}$.
As detailed in Sec.~\ref{sec:spec_loss}, we select the pair of adverse and normal patch embeddings, which surpass a warping confidence threshold, denoted as $\textbf{f}_\textrm{adv}^{i}$ and ${\textbf{c}}_\textrm{norm}^{i}$ for the $i^\textrm{th}$ patch embedding.
To guide the condition of our segmentation feature $\textbf{f}_\textrm{adv}^{i}$ to resemble the normal condition while only considering condition information, we project our feature $\textbf{f}_\textrm{adv}^{i}$ and normal condition-infused feature ${\textbf{c}}_{\textrm{norm}}^{i}$ on the condition embedding space, \ie, $\psi_\textrm{proj}(\textbf{f}_{\textrm{adv}}^{i})$ and ${\textbf{z}}_\textrm{norm}^{i}=\psi_\textrm{proj}({\textbf{c}}_{\textrm{norm}}^{i})$, respectively.
Then, we approximate $\psi_\textrm{proj}(\textbf{f}_{\textrm{adv}}^{i})$ to ${\textbf{z}}_\textrm{norm}^{i}$ using an $\ell$1 regression loss as follows:
\begin{align}
\mathcal{L}_{\textrm{resto}} = \frac{1}{|\mathcal{W}|}\sum\limits_{i \in \mathcal{W}}|\psi_\textrm{proj}(\textbf{f}_{\textrm{adv}}^{i})-{\textbf{z}}_\textrm{norm}^{i}|,
\label{eq:loss_norm_mimicking}
\end{align}
where $\mathcal{W} = \{i~|~\mathrm{conf}(i)\geq0.2\}$ with $\mathrm{conf}(i)$ denoting the warping confidence.

\vspace{2mm}\noindent\textbf{Adverse Condition Discriminating Loss.}
To further facilitate feature restoration, we present another loss that pushes the encoder feature and condition-infused feature of an adverse condition image apart.
To this end, we introduce an MLP-based adverse condition discriminator denoted as $D$.
It discriminates the encoder feature $\textbf{f}_\textrm{adv}^{l,j}$ and condition-infused feature $\textbf{c}_\textrm{adv}^{l,j}$ for each $l^\textrm{th}$ layer and $j^\textrm{th}$ patch embedding for $j \in \mathcal{A}$ where $\mathcal{A}$ is a set of indices of all patch embeddings.
Then, $\textbf{f}_\textrm{adv}^{l,j}$ and $\textbf{c}_\textrm{adv}^{l,j}$ are vectorized and forwarded to the discriminator, which is trained with the cross-entropy loss, denoted as $\mathcal{L}_{\textrm{dis}}$, to classify them into two classes: encoder feature and condition-infused feature.
The detailed architecture of $D$ is shown in \Fig{loss_dis} and the loss is given by
\begin{align}
\mathcal{L}_{\textrm{dis}} = - \frac{1}{|\mathcal{A}|}\displaystyle\sum_{j \in \mathcal{A}}\sum_{l=1}^{L}\big\{\lambda\log(D(\textbf{f}_\textrm{adv}^{l,j}))+(1-\lambda)\log(1-D(\textbf{c}_\textrm{adv}^{l,j}))\big\},
\label{eq:loss_adv_discriminating}
\end{align}
where $L$ denotes the number of layers of the encoder, and $\lambda=0$ if the input is the adverse condition-infused feature $\textbf{c}_\textrm{adv}^l$ and $\lambda=1$ otherwise.

\vspace{2mm}\noindent\textbf{Total Loss for Step 2.}
Besides the proposed losses, we utilize two conventionally used loss functions: one for self-training $\mathcal{L}_\textrm{self}$ and the other for entropy minimization $\mathcal{L}_\textrm{ent}$, following the previous work~\cite{bruggemann2023contrastive}. 
The segmentation network is trained by minimizing 
$\mathcal{L}_\textrm{step2} = \mathcal{L}_\textrm{resto} + \lambda_\textrm{dis}\mathcal{L}_\textrm{dis} + \mathcal{L}_\textrm{self} + \lambda_\textrm{ent}\mathcal{L}_\textrm{ent} $.

\setlength{\floatsep}{10pt plus 0pt minus 5pt}
\setlength{\textfloatsep}{10pt plus 0pt minus 5pt}
\begin{figure}[t!]
    \centering
    \scalebox{1.0}{
    \includegraphics[width=\linewidth]{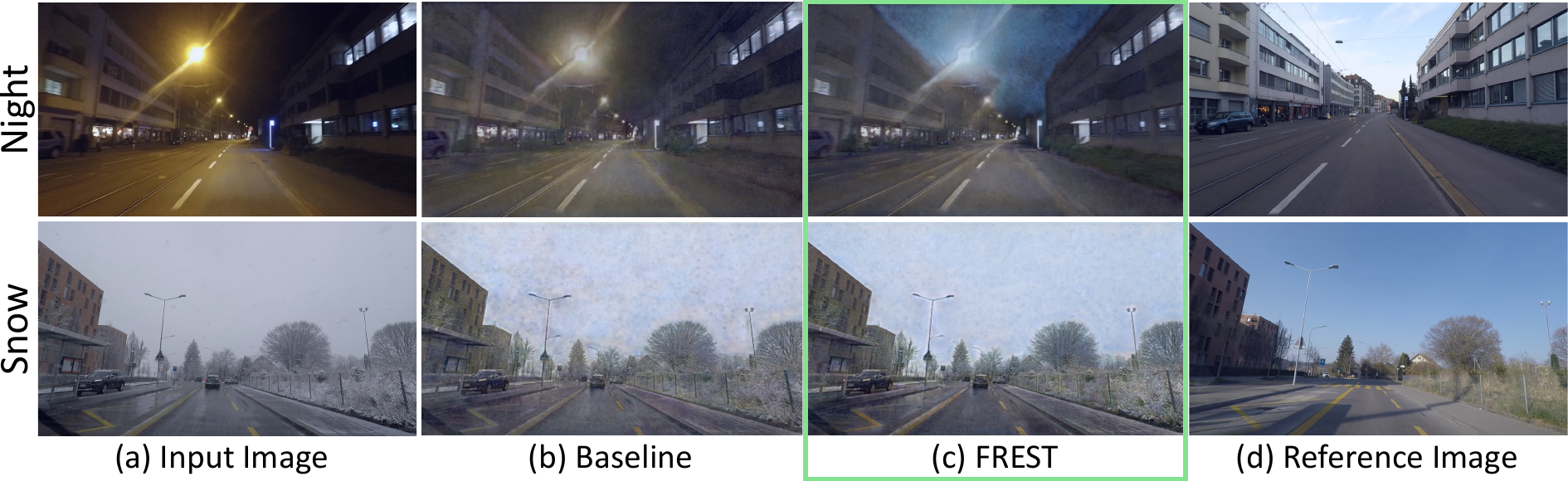}}
    \caption{Image reconstruction results from segmentation features. (a) Input target image. Image reconstructed by the (b) Baseline~\cite{xie2021segformer} and (c) FREST. (d) Reference image.}
    \label{fig:recon}
\end{figure}

\setlength{\floatsep}{10pt plus 0pt minus 5pt}
\setlength{\textfloatsep}{10pt plus 0pt minus 5pt}
\begin{figure}[t!]
    \centering
    \scalebox{1.0}{
    \includegraphics[width=\linewidth]{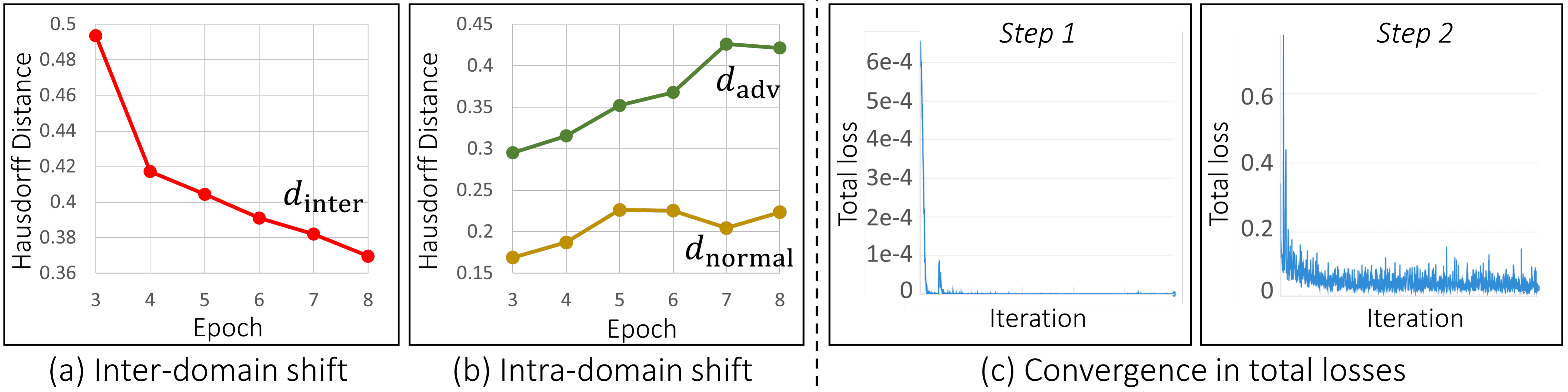}}
    \caption{Empirical analysis on the impact of feature restoration during training FREST. (a) Inter-domain shift between adverse and normal conditions. (b) Intra-domain shift within each condition. (c) Convergence in total losses for both Step 1 and Step 2.}
    \label{fig:quant}
\end{figure}

\subsection{Empirical Justification}\label{sec:empirical}
To investigate the effect of FREST, we first conduct a qualitative analysis by reconstructing images from the restored features learned by FREST.
Please note that FREST learns feature restoration for robustly recognizing adverse condition images as if they were in normal condition, does not image reconstruction during both training and testing.
For image reconstruction, we adopt SegFormer~\cite{xie2021segformer} as a reconstruction model with upsampling layers.
The decoder of the reconstruction model is trained for reconstruction on normal images while the encoder is pretrained on Cityscapes~\cite{cityscapes} and then frozen.
The encoder is then replaced with that of FREST ($\phi_\textrm{enc}$).
We also conduct the reconstruction using a baseline~\cite{xie2021segformer} in the same manner.
\Fig{recon} shows the favorable impact of FREST: a night sky turns blue, and a snowy tree appears green.

The impact of feature restoration is also demonstrated by the quantitative analysis in \Fig{quant}.
We measure the inter-domain shift in \Fig{quant}(a), the distance between feature distributions of adverse and normal domains ($d_{\text{inter}}$), and the intra-domain shift in \Fig{quant}(b), the distance between feature distributions at before training and after $n$ epochs of training with FREST, for each domain ($d_{\text{adv}}$ and $d_{\text{normal}}$). 
We adopt the Hausdorff distance~\cite{huttenlocher1993comparing} using cosine distance to measure all such shifts.
Our analysis reveals that FREST restores adverse features as desired: the adverse feature distribution gradually approaches to the normal feature distribution during training, while the normal feature distribution changes little.

As shown \Fig{quant}(c), we investigate the convergence in total losses for both Step 1 and Step 2 during training FREST.
The results demonstrate that FREST successfully converges during training while the two steps are performed in each iteration alternatingly.
Since $\mathcal{L}_\textrm{spec}$ and $\mathcal{L}_\textrm{resto}$ have opposite objectives conducted in condition embedding space, it might seem optimizing two losses would hinder the convergence of FREST.
However, this issue does not occur since we separate the trainable parameters for each step: in Step 1, we train the condition strainer and projection head, while in Step 2, we train the segmentation network.
This training strategy ensures that each step is optimized independently, without conflict, enabling FREST to converge as intended.


\section{Experiments}
\label{sec:experiments}
\subsection{Experimental Setting}

\noindent\textbf{Datasets.}
For our experiments, we utilize Cityscapes~\cite{cityscapes} as the source domain, and we use ACDC~\cite{acdc} and RobotCar Correspondence~\cite{larsson2019cross,maddern20171} as the target domains, respectively.
We also use the ACG Benchmark~\cite{bruggemann2023contrastive} for evaluating the generalization capability of our model on diverse adverse conditions.
Lastly, we evaluate our method in the normal condition on Cityscapes-lindau40, a dataset commonly used in robust visual recognition research~\cite{dai2020curriculum,lee2022fifo}.

\noindent\textbf{Implementation Details.}
We adopt SegFormer~\cite{xie2021segformer} architecture as our segmentation network, which is pre-trained on the Cityscapes dataset~\cite{cityscapes}.
The segmentation network is trained by an AdamW~\cite{loshchilov2017decoupled} optimizer with weight decay $\expnum{1}{2}$.
The initial learning rate is set to $\expnum{1}{5}$ for the encoder and decoder of the segmentation network and $\expnum{5}{4}$ for the condition strainer.
In addition, we employ a linear learning rate decay with a linear warm-up for the first 1,500 iterations.
Finally, the hyper-parameters are set to $\lambda_\textrm{spec}$, $\lambda_\textrm{ent}$, $\lambda_\textrm{dis}$, and $\tau$ as $\expnum{1}{2}$, $\expnum{1}{2}$, $\expnum{5}{5}$, and $\expnum{7}{1}$, respectively.
More details are given in the supplement.

\setlength{\floatsep}{10pt plus 0pt minus 5pt}
\setlength{\textfloatsep}{10pt plus 0pt minus 5pt}
\begin{table*}[t!]
\smallskip
\centering
\caption{Comparison with existing methods on Cityscapes $\to$ ACDC.
The results are reported in mIoU (\%) on the ACDC test set.}%
\scalebox{0.62}{
\ra{1.3}
\begin{tabular}{@{}lcccccccccccccccccccr@{}}\toprule
\multirow{3}{*}{Method} & \multicolumn{19}{c}{ACDC IoU} \\
\cmidrule{2-21} & \rotatebox[origin=c]{90}{road} & \rotatebox[origin=c]{90}{sidew.} & \rotatebox[origin=c]{90}{build.} & \rotatebox[origin=c]{90}{wall} & \rotatebox[origin=c]{90}{fence} & \rotatebox[origin=c]{90}{pole} & \rotatebox[origin=c]{90}{light} & \rotatebox[origin=c]{90}{sign} & \rotatebox[origin=c]{90}{veget.} & \rotatebox[origin=c]{90}{terrain} & \rotatebox[origin=c]{90}{sky} & \rotatebox[origin=c]{90}{person} & \rotatebox[origin=c]{90}{rider} & \rotatebox[origin=c]{90}{car} & \rotatebox[origin=c]{90}{truck} & \rotatebox[origin=c]{90}{bus} & \rotatebox[origin=c]{90}{train} & \rotatebox[origin=c]{90}{motorc.} & \rotatebox[origin=c]{90}{bicycle} & \multicolumn{1}{c}{\phantom{00}\rotatebox[origin=c]{90}{\textbf{mean}}} \\ \midrule
Source model~\cite{xie2021segformer}  & 85.7 & 51.0 & 76.6 & 36.4 & 37.1 & 45.2 & 55.7 & 57.5 & 77.7 & 52.0 & 84.1 & 60.3 & 34.8 & 82.9 & 61.6 & 65.4 & 73.4 & 37.9 & 52.5 & 59.4 \\
HCL~\cite{huang2021model} & 86.4 & 53.5 & 78.5 & 38.8 & 38.1 & 48.0 & 57.8 & 58.9 & 78.1 & 52.4 & 85.1 & 61.7 & 37.1 & 83.7 & 64.1 & 66.6 & 74.5 & 39.1 & 53.3 & 60.8 \\
URMA~\cite{fleuret2021uncertainty} & 89.2 & 60.4 & 84.3 & 48.7 & 42.5 & 53.8 & 65.4 & 63.8 & 76.3 & 57.3 & 85.9 & 63.4 & 43.9 & 85.8 & \underline{68.8} & 73.2 & 82.8 & 46.3 & 48.4 & 65.3 \\
URMA + SimT~\cite{guo2022simt} & 90.0 & 65.7 & 80.6 & 46.0 & 41.7 & \underline{56.3} & 65.2 & 62.7 & 75.9 & 55.6 & 84.4 & 66.4 & \underline{46.6} & 85.4 & 68.4 & 72.3 & 80.0 & \underline{46.8} & 58.0 & 65.7 \\
CMA~\cite{bruggemann2023contrastive} & \textbf{94.0} & \textbf{75.2} & \textbf{88.6} & \underline{50.5} & \underline{45.5} & 54.9 & \underline{65.7} & \underline{64.2} & \textbf{87.1} & \textbf{61.3} & \underline{95.2} & \underline{67.0} & 45.2 & \underline{86.2} & 68.6 & \underline{76.6} & \textbf{83.9} & 43.3 & \underline{60.5} & \underline{69.1} \\
FREST (Ours) & \underline{93.3} & \underline{72.2} & \underline{88.3} & \textbf{52.4} & \textbf{46.6} & \textbf{58.6} & \textbf{66.2} & \textbf{66.1} & \underline{86.1} & \underline{58.6} & \textbf{95.3} & \textbf{69.9} & \textbf{49.2} & \textbf{89.1} & \textbf{75.1} & \textbf{79.4} & \underline{83.0} & \textbf{52.9} & \textbf{61.4}  & \textbf{70.7} \\
\bottomrule
\end{tabular}
}
\label{tab:acdc_sota}
\end{table*}

\setlength{\floatsep}{10pt plus 0pt minus 5pt}
\setlength{\textfloatsep}{10pt plus 0pt minus 5pt}
\begin{table}[t!]
\centering
  \begin{minipage}{0.23\linewidth}
    \centering
    \caption{Comparison with previous methods on City $\to$ RobotCar.
     }    
    \scalebox{0.73}{\begin{tabular}{@{}lc@{}}\toprule
    {Method} & mIoU \\ 
    \midrule
    Source model~\cite{xie2021segformer} & 50.0  \\
    HCL~\cite{huang2021model} & 50.1   \\
    URMA~\cite{fleuret2021uncertainty} & 51.6  \\
    URMA + SimT~\cite{guo2022simt} & 52.4  \\
    CMA~\cite{bruggemann2023contrastive} & \underline{54.3}  \\
    FREST (Ours) & \textbf{58.8} \\
    \bottomrule
    \end{tabular}}
    \label{tab:comp_robotcar}
  \end{minipage}%
  \quad 
  \begin{minipage}{0.26\linewidth}
    \centering
    \caption{
    Comparison with UDA methods on City $\to$ ACDC.
    ($\mathcal{SF}$: source-free method).
    }
    \ra{1.06}
    \scalebox{0.7}{
    \begin{tabular}{@{}lcc@{}}\toprule
    Method & $\mathcal{SF}$ & mIoU \\
    \midrule
    Source model~\cite{xie2021segformer} & & 59.4  \\
    Refign~\cite{bruggemann2023refign} & & 65.5 \\
    HRDA~\cite{hoyer2022hrda} & & 68.0 \\
    HRDA + MIC~\cite{hoyer2023mic}  &  &  \underline{70.4}  \\
    CMA~\cite{bruggemann2023contrastive} & \checkmark & 69.1 \\
    FREST (Ours) & \checkmark & \textbf{70.7} \\
    \bottomrule
    \end{tabular}
    }
    \label{tab:comp_uda}
  \end{minipage}
\quad 
\begin{minipage}{0.44\linewidth}
    \centering  
    \caption{
    Generalization performance of models adapted from Cityscapes to ACDC on ACG and Cityscapes-lindau40 (C-Lindau).
    }
    \ra{1.08}
    \scalebox{0.65}{
    \begin{tabular}{@{}lccccccccr@{}}\toprule
\multirow{2}{*}{Method}  && \multicolumn{5}{c}{ACG} & C-Lindau\\
\cmidrule(lr){3-7} \cmidrule(lr){8-8}
&& fog & night & rain & snow & \textbf{all} & normal\\
\midrule
SegFormer~\cite{xie2021segformer}  && 54.0 & 27.9 & 47.5 & 41.2 & 40.1 & \textbf{72.7}\\
HCL~\cite{huang2021model}  && 54.2 & 28.3 & 48.2 & 42.4 & 40.8 & - \\
URMA~\cite{fleuret2021uncertainty}  && 54.1 & 31.0 & 51.9 & 45.5 & 44.4 & - \\
CMA~\cite{bruggemann2023contrastive} && 59.7 & \textbf{40.0} & \underline{59.6} & \textbf{52.2} & \underline{51.3} & 71.8\\
FREST (Ours) && \textbf{61.4} & \underline{39.9} & \textbf{61.0} & \underline{51.9} & \textbf{52.6} & \underline{72.5}\\
\bottomrule
\end{tabular}
    }
\label{tab:comp_acg}
  \end{minipage}
\end{table}

\subsection{Quantitative Results}
We first compare FREST with existing methods for SFDA under multiple adverse conditions on Cityscapes $\to$ ACDC and Cityscapes $\to$ RobotCar benchmarks.
We extend our comparison to the previous UDA methods on Cityscapes $\to$ ACDC benchmark.
Lastly, we evaluate the generalization capability of FREST on ACG and Cityscapes-lindau40 compared with existing SFDA methods.

\noindent\textbf{Comparison with SFDA Methods.}
As summarized in~\Tbl{acdc_sota}, FREST achieves state-of-the-art performance with a notable improvement of 1.6\% in mIoU over the most recent work~\cite{bruggemann2023contrastive} on Cityscapes $\to$ ACDC benchmark, especially improving on fine-grained objects (\eg, car, truck, bus) by large margins.
As shown in~\Tbl{comp_robotcar}, FREST substantially outperforms all previous SFDA models on Cityscapes $\to$ RobotCar benchmark.
Considering that the RobotCar dataset includes a wider range of conditions (\ie, dawn, dusk, night, night-rain, overcast, rain, snow, and sun) compared to the ACDC dataset, which has only four conditions (\ie, fog, rain, snow, and night), these results demonstrate that our model becomes increasingly effective in improving robustness as the diversity of adverse conditions increases.

\noindent\textbf{Comparison with UDA Methods.}
As summarized in~\Tbl{comp_uda}, FREST outperforms the existing UDA methods on Cityscapes $\to$ ACDC.
Note that UDA methods utilize both a labeled source dataset and an unlabeled target dataset during adaptation, while SFDA methods including FREST use only the unlabeled target dataset.
The results show that our framework surpasses the UDA methods, even without access to a labeled source domain during the adaptation.

\begin{table}[t!]
\centering
  \begin{minipage}{0.43\linewidth}
    \caption{
    Loss analysis on (a) Step 1 and (b) Step 2.
    }
  \renewcommand{\arraystretch}{1.1}
    \centering
    \begin{subtable}[h]{0.45\linewidth}
    \scalebox{0.92}{
    \begin{tabular}{@{}ccc@{}}\toprule
     $\mathcal{L}_\textrm{self}$     &  	 $\mathcal{L}_\textrm{spec}$  & mIoU \\ 
    \midrule
      &  &    64.3    \\
     \checkmark &  &   64.8  \\
     \checkmark & \checkmark  &  \textbf{68.6}       \\
    \bottomrule
    \end{tabular}
    }
    \caption{}
    \end{subtable}
    \begin{subtable}[h]{0.45\linewidth}
    \scalebox{0.92}{
    \begin{tabular}{@{}ccc@{}}\toprule
    $\mathcal{L}_\textrm{resto}$   &  $\mathcal{L}_\textrm{dis}$ & mIoU \\ 
    \midrule
           &    &   62.7  \\
          \checkmark &    &   67.2   \\
          \checkmark &  \checkmark   &       \textbf{68.6} \\
    \bottomrule
    \end{tabular}
    }
    \caption{}
    \end{subtable}
    \label{tab:abl_loss}
  \end{minipage}
  \quad 
  \begin{minipage}{0.52\linewidth}
    \centering
    \caption{Analysis of the structure and training strategy in FREST. 
     }    
    \scalebox{0.88}{\begin{tabular}{ccccc}
    \toprule
    \multicolumn{2}{c}{Trainable Module } & \multicolumn{2}{c}{Training Strategy}  & \multirow{2}{*}{ mIoU} \\
    \cmidrule(lr){1-2} \cmidrule(lr){3-4}
    Strainer  &  Seg. &  Self-training&FREST  & \\ 
    \midrule
      &  \checkmark    & \checkmark  &    &    62.7    \\
    \checkmark  &  &     \checkmark  &        &   63.1  \\
    \checkmark  &  \checkmark &     \checkmark  &        &   63.2  \\
     \checkmark  &   \checkmark &           & \checkmark & \textbf{68.6} \\
     \bottomrule
    \end{tabular}}
    \label{tab:abl_comp}
  \end{minipage}
\end{table}

\noindent\textbf{Generalization Capability.}
Following~\cite{bruggemann2023contrastive}, we evaluate the generalization capabilities of FREST and the competitors, adapted from Cityscapes to ACDC, on the ACG benchmark containing multiple adverse conditions (\ie, fog, night, rain, and snow).
As shown in \Tbl{comp_acg}, FREST outperforms the previous methods for all ACG samples, which shows the robust generalizability of FREST in adverse conditions.
Furthermore, we extend our evaluation to the Cityscapes-lindau40 dataset to investigate the generalizability under normal conditions.
FREST surpasses CMA~\cite{bruggemann2023contrastive} and performs on par with SegFormer~\cite{xie2021segformer} which is trained on Cityspaces as a labeled source dataset.
The results indicate that FREST effectively generalizes on normal conditions as well as adverse conditions by its feature restoration, converting features from adverse to normal conditions.

\subsection{Ablation Study}

We first investigate the effect of each loss by the ablation study in~\Tbl{abl_loss}.
The results show that all the losses contribute to the performance in each step.
In particular, the condition-specific loss in Step 1 and the feature restoration loss in Step 2 significantly fulfill the primary objectives of FREST, contributing significantly to improving the robustness of semantic segmentation.

In~\Tbl{abl_comp}, we analyze the effect of the condition strainer, which draws its structural inspiration from the adapter structure~\cite{houlsby2019parameter} for parameter-efficient learning.
To this end, we conduct a comparative study of FREST against the naive fine-tuning strategy using the condition strainer (\ie, adapter).
As shown in the second row of the table, the conventional fine-tuning scheme using the adapter improves the performance marginally, demonstrating the ineffectiveness of a naive application of the adapter (+0.4\%p). 
In addition, the third row shows that the full fine-tuning of all parameters of both the segmentation network and the adapter results in a slight improvement (+0.5\%p).
It suggests that our performance improvement does not solely stem from an increase in the number of parameters in the condition strainer. 
Consequently, the proposed training scheme, coupled with the condition strainer, contributes to the performance significantly (+5.9\%p).

\begin{table}[t!]
\centering
  \begin{minipage}{0.48\linewidth}
  \centering
  \begin{minipage}{\linewidth}
  \centering
    \caption{The number of parameters for additional modules.
    }
    \renewcommand{\arraystretch}{0.9}
  \scalebox{0.9}{
    \begin{tabular}{lc}
    \toprule
       & Param. \\ \midrule
    SegFormer~\cite{xie2021segformer}       & 81.4M (100\%)\\
    Condition Strainer      &  2.1M (2.6\%)\\
    Projection Layer  &  1.2M (1.5\%)\\ 
    \bottomrule
    \end{tabular}}\label{tab:analy_param}
    \end{minipage}
  \begin{minipage}{\linewidth}
  \centering
    \caption{Impact of restored features $\textbf{f}_\textrm{adv}$.
    }
    \renewcommand{\arraystretch}{0.9}
\scalebox{0.9}{\begin{tabular}{lc}
    \toprule
    Features for Inference & mIoU \\ 
    \midrule
    Condition-infused feature $\textbf{c}_\textrm{adv}$   &  59.0   \\
     Restored feature $\textbf{f}_\textrm{adv}$    &  \textbf{68.6}   \\
     \bottomrule
    \end{tabular}}
    \label{tab:condition_specific_inference}
    \end{minipage}
  \end{minipage}
  \quad 
  \begin{minipage}{0.48\linewidth}
    \centering
    \caption{
    Performance according to positive embedding selection strategies and loss functions. \textit{Cls.} and \textit{Contra.} denote classification and contrastive loss.
    }
    \renewcommand{\arraystretch}{1.33}
    \scalebox{0.81}{
    \begin{tabular}{l|cc|c}
    \toprule
    Pos. Emb. Selection    & ~\textit{Cls.}~  &  ~\textit{Contra.}~    & ~mIoU~  \\ \midrule
    \multirow{2}{*}{All}           & \checkmark     &                      & 62.9  \\
                                   &                &  \checkmark          & 63.4  \\ \midrule
    (1) $\mathtt{RANDOM}$          &                &  \checkmark          & 62.4 \\
    (2) $\mathtt{LOWEST}$          &                &  \checkmark          &  56.6 \\
    (3) $\mathtt{HIGHEST}$ (Ours)  &                &  \checkmark          &  \textbf{68.6}  \\ 
    \bottomrule
    \end{tabular}
    }
    \label{tab:analy_csl}
  \end{minipage}
\end{table}

\subsection{Analysis on FREST}\label{sec:patch}

\noindent\textbf{Parameter Efficiency.}
As shown in \Tbl{analy_param}, the total parameters of the baseline, SegFormer~\cite{xie2021segformer}, are 81.4M (100\%), while the strainers occupy only 2.1M (2.6\%), and the projection head has 1.2M (1.5\%), which require only a small number of parameters.
Also, it's important to note that no additional parameters are required during inference, as only the encoder and decoder are utilized.

\noindent\textbf{Impact of Restored Feature $\textbf{f}_\textrm{adv}$.}
We evaluate the efficacy of the restored adverse features $\textbf{f}_\textrm{adv}$ learned by FREST.
To this end, we compare the inference result using the restored features $\textbf{f}_\textrm{adv}$, as employed in our framework, with that using the condition-infused features $\textbf{c}_\textrm{adv}$.
\Tbl{condition_specific_inference} demonstrates that utilizing restored features $\textbf{f}_\textrm{adv}$ robustly performs as intended, while employing condition-infused features $\textbf{c}_\textrm{adv}$ leads to inferior performance due to their inclusion of detrimental characteristics of adverse conditions.

\noindent\textbf{Analysis on Condition-specific Learning.}
To verify our design choice for the objective function learning condition information, we investigate variants of its positive embedding selection strategies and loss functions as summarized in~\Tbl{analy_csl}.
In detail, we first evaluate variants using all condition embeddings $\{\mathbf{z}^{i}_{\textrm{adv}}\}^N_{i=0}$ where $N$ is the number of condition embeddings in a positive queue varying classification and contrastive loss as loss functions.
We adopt supervised contrastive learning~\cite{khosla2020supervised} for multiple positive samples.
The results show that the contrastive loss learns better condition embedding space than the classification loss.
To address the diverse distribution of positive embeddings from adverse conditions in contrastive learning, we choose the representative positive embedding following variants: (1) $\mathtt{RANDOM}$ that select an arbitrary embedding in the positive queue, (2) $\mathtt{LOWEST}$ that pick the lowest similar embedding with the anchor embedding, and (3) $\mathtt{HIGHEST}$ that pick the highest similar embedding.
The results demonstrate that using the highest similar embedding as a positive sample for contrastive learning is the most effective strategy for learning condition embedding space.
We suspect the reason is that the most similar positive embedding is likely to share common adverse conditions with the anchor embedding.
This similarity helps to learn condition information, allowing the model to consider the distinct attributes of each adverse condition. Consequently, it helps FREST to learn the condition embedding space effectively.

\begin{figure*}[t]
    \centering
    \includegraphics[width=\linewidth]{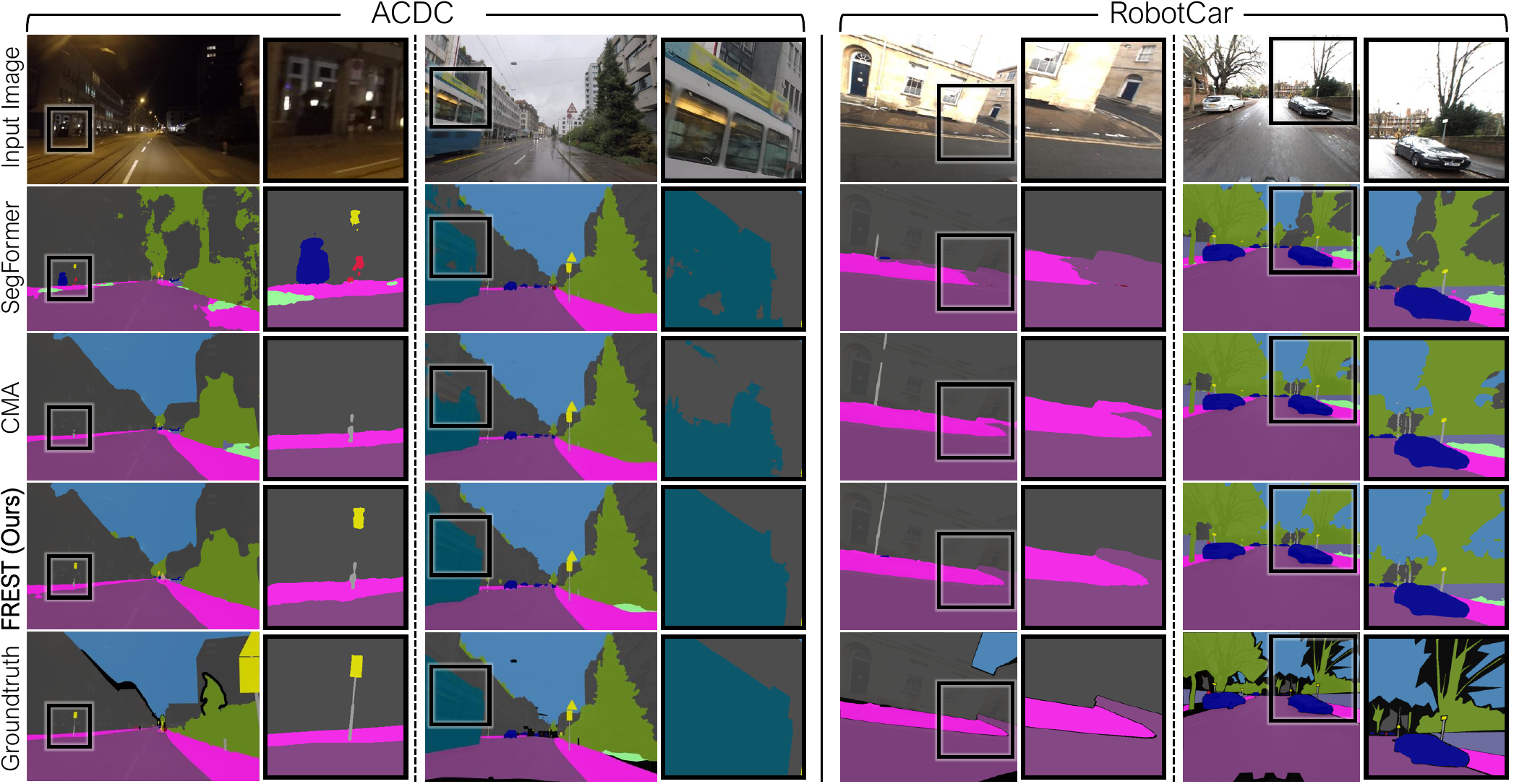}
    \caption{
    Qualitative results of FREST (Ours), its baseline (SegFormer~\cite{xie2021segformer}), and CMA~\cite{bruggemann2023contrastive} on ACDC and RobotCar.
    }
\label{fig:qual_main}
\end{figure*}

\subsection{Qualitative Results}
As illustrated in~\Fig{qual_main}, we present the qualitative results FREST, its baseline~\cite{xie2021segformer}, and a previous work~\cite{bruggemann2023contrastive} on ACDC~\cite{acdc} and RobotCar~\cite{larsson2019cross,maddern20171}.
The results show that FREST excels in segmenting fine-grained objects such as a pole (1st and 4th column) and classifying ambiguous semantics such as road and sidewalk (2nd and 3rd column) across multiple adverse conditions compared with its baseline and the previous work.

\section{Conclusion}\label{conclusion}
We have presented the novel framework of feature restoration for multiple adverse conditions.
FREST operates in two stages as follows: (1) learning the condition embedding space that represents only condition-specific information of images and (2) restoring features for adverse conditions on the learned condition embedding space.
As a result, FREST achieved a new state of the art on two benchmarks for SFDA under multiple adverse conditions, while it showed superior generalization ability on unseen datasets.
In terms of limitations, our framework currently does not cover various adverse conditions, including image degradation and camera artifacts, which will be expanded in our future work.

{\small
\noindent \textbf{Acknowledgement.} 
This work was supported by Samsung Research Funding \& Incubation Center of Samsung Electronics under Project Number SRFC-IT1801-52.
}
\pagebreak

\appendix
\setcounter{table}{0}
\setcounter{figure}{0}
\setcounter{equation}{0}
\renewcommand{\theequation}{a\arabic{equation}}
\renewcommand{\thetable}{a\arabic{table}}
\renewcommand{\thefigure}{a\arabic{figure}}

This supplementary material presents additional experimental details and results that are omitted from the main paper due to the space limit.
Firstly, we include more implementation details in Sec.~\ref{appendix:details}.
Then, Sec.~\ref{appendix:algorithm} shows the algorithm of FREST and Sec.~\ref{appendix:empirical_analysis} presents a thorough analysis of FREST, covering aspects such as the analysis on FREST via \textit{t}-SNE and the impact of the adverse condition discriminating loss.
Finally, we offer an additional ablation study in Sec.~\ref{appendix:ablation}, containing combinations of losses, patch confidence threshold, and extended quantitative and qualitative results in Sec.~\ref{appendix:conditionwise} and Sec.~\ref{appendix:additional_qual}.

\section{Implementation Details}\label{appendix:details}

In this section, we present the implementation settings that are omitted from the main paper.
All experiments were conducted using a single A6000 GPU.
We train both the segmentation network and the condition strainer for 8 epochs while training only the condition strainer for 2 epochs to train a stable condition strainer.
Each mini-batch consists of one image for each adverse and normal image.
During training, images are cropped to 1080$\times$1080 and flipped horizontally at random.
Additionally, we utilize the exponential moving average (EMA) model for implementing a model with a stopping gradient in our framework, the weight parameter is set as 0.9999 to preserve the source knowledge.
For the condition strainer, we initialize up-projecting parameters with He uniform initialization and down-projecting parameters with zero initialization.
This initialization strategy aims to optimize the learning efficiency and model stability in handling condition-specific features.

\section{Algorithm of FREST}\label{appendix:algorithm}
We present the training procedure of FREST in Algorithm~\ref{alg:alg}.

\begin{algorithm} [ht]
    \caption{: Training Algorithm of FREST}\label{alg:alg}
    \textbf{Input:} Condition strainer: $\psi_\textrm{strainer}$, Projection head: $\psi_\textrm{proj}$, $l^\textrm{th}$ layer of encoder: $\phi_{\textrm{enc}}^l$, Decoder: $\phi_{\textrm{dec}}$, Number of layers: $L$, Prediction: $P$, Pseudo Label: $\hat{Y}$, Input images: $\{x_\textrm{adv}$, $x_\textrm{norm}\}$
    
    \textbf{Output:} Optimized encoder $\phi_{\textrm{enc}}$ and decoder  $\phi_{\textrm{dec}}$.
    
    \begin{algorithmic}[1]
    
    \For{\{1, \dots , \# of training iterations\}}
        \State Sample mini-batch $\{ x_\textrm{adv}, x_\textrm{norm} \}$
        \For{\{{$l \longleftarrow 1$ to $L$}\}}
        \State  $\textbf{c}_\textrm{adv}^{l} = \phi_{\textrm{enc}}^l(\textbf{c}_\textrm{adv}^{l-1})+\psi_\textrm{strainer}^l(\textbf{c}_\textrm{adv}^{l-1})$ 
        \State  $\textbf{c}_\textrm{norm}^{l} = \phi_{\textrm{enc}}^l(\textbf{c}_\textrm{norm}^{l-1})+\psi_\textrm{strainer}^l(\textbf{c}_\textrm{norm}^{l-1})$
        \EndFor
        \State $\mathcal{L}_\textrm{spec} = \mathcal{L}_{\textrm{spec}}(\psi_\textrm{proj}(\textbf{c}_\textrm{adv}), \psi_\textrm{proj}({\textbf{c}}_\textrm{norm}))$ \Comment{Eq.~(1)\&(2)}
        \State \textbf{Update} projection head $\psi_\textrm{proj} $ 
        \State $\mathcal{L}_\textrm{self} = \mathcal{L}_\textrm{self}(P_\textrm{adv},\hat{Y}_\textrm{adv})$ 
        \State $\mathcal{L}_{\textrm{strainer}} = \lambda_\textrm{spec}\mathcal{L}_{\textrm{spec}} + \mathcal{L}_\textrm{self}$
        \State \textbf{Update} condition strainer $\psi_\textrm{strainer} $ \textrm{with} $\mathcal{L}_{\textrm{strainer}}$ \Comment{Step 1}
        \For{\{{$l \longleftarrow 1$ to $L$}\}}
        \State  $\textbf{f}_\textrm{adv}^{l} = \phi^l_{\textrm{enc}}(\textbf{f}_\textrm{adv}^{l-1})$
        \State  $\textbf{c}_\textrm{adv}^{l} = \phi_{\textrm{enc}}^l(\textbf{f}_\textrm{adv}^{l-1})+\psi_\textrm{strainer}^l(\textbf{f}_\textrm{adv}^{l-1})$ 
        \State $\mathcal{L}_\textrm{dis}^l = -\mathcal{L}_\textrm{dis}^l(\textbf{f}_\textrm{adv}^l,\textbf{c}_\textrm{adv}^l)$ \Comment{Eq.~(4)}
        \EndFor
        \State $\mathcal{L}_\textrm{self} = \mathcal{L}_\textrm{self}(P_\textrm{adv},\hat{Y}_\textrm{adv})$ 
        \State $\mathcal{L}_\textrm{resto} = \mathcal{L}_\textrm{resto}(\psi_\textrm{proj}(\textbf{f}_\textrm{adv}),\psi_\textrm{proj}({\textbf{c}}_\textrm{norm}))$ \Comment{Eq.~(3)}
        \State $\mathcal{L}_{\textrm{total}} =
        \mathcal{L}_\textrm{self} + \lambda_\textrm{ent}\mathcal{L}_\textrm{ent} + \mathcal{L}_\textrm{resto} + \lambda_\textrm{dis}\sum_{l=1}\mathcal{L}_\textrm{dis}^l$
        \State \textbf{Update} $\phi_{\textrm{enc}}$ and $\phi_{\textrm{dec}}$ of the model \textrm{with} $\mathcal{L}_{\textrm{total}}$ 
    \Comment{Step 2}
    \EndFor
    
    \end{algorithmic}
\end{algorithm}\label{algo}

\section{Empirical Analysis}\label{appendix:empirical_analysis}

\subsection{Analysis on FREST}
\begin{figure}[h]
    \centering
    \vspace{0mm}
    \scalebox{1.4}{
    \includegraphics[width=0.5\linewidth]{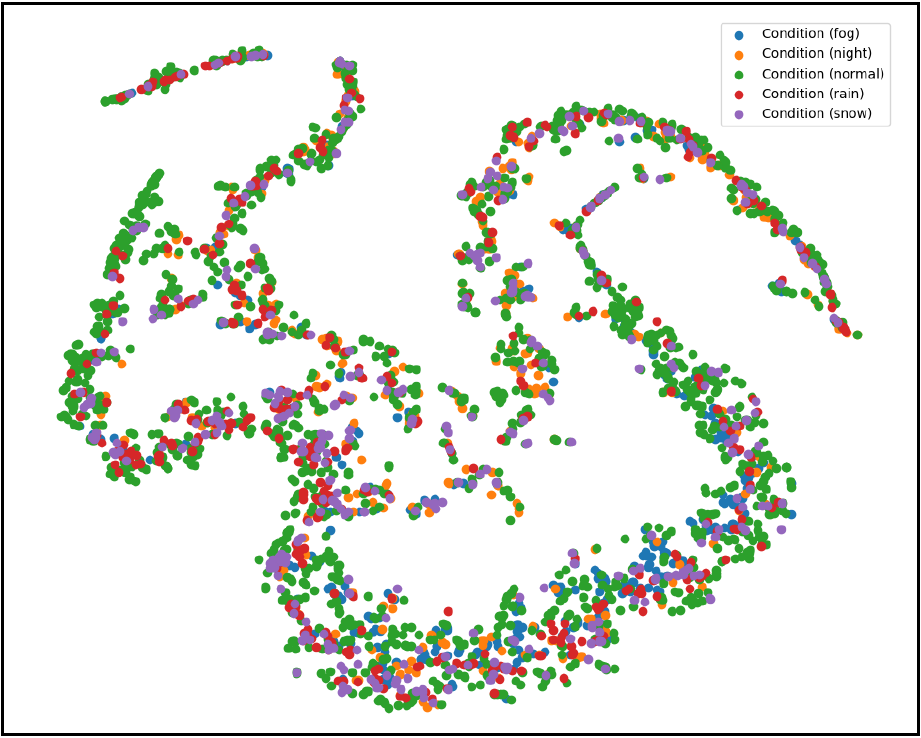}}
\caption{
\textit{t}-SNE visualization of the distribution of condition embeddings in the condition embedding space.
} \label{fig:tsne}
\end{figure}

To investigate the impact of FREST, we show that it reduces the condition gaps between adverse and normal conditions.
To this end, visualize \textit{t}-SNE~\cite{van2008visualizing} using our segmentation features under adverse conditions (\ie~fog, night, rain, and snow) and condition-specific features under the normal condition (\ie~normal) in the condition embedding space learned by FREST.
In detail, we utilize the validation images from the ACDC dataset~\cite{acdc} input images, and we compute their condition embeddings using both the condition strainer and the projection head.
\Fig{tsne} shows that FREST effectively reduces the condition gaps between adverse and normal conditions well, which suggests that FREST achieves condition-invariance through feature restoration.

\subsection{Additional Analysis on Fig.~6}\label{appendix:analysis_fig6}
\begin{figure}[h]
    \centering
    \scalebox{0.98}{
    \includegraphics[width=\linewidth]{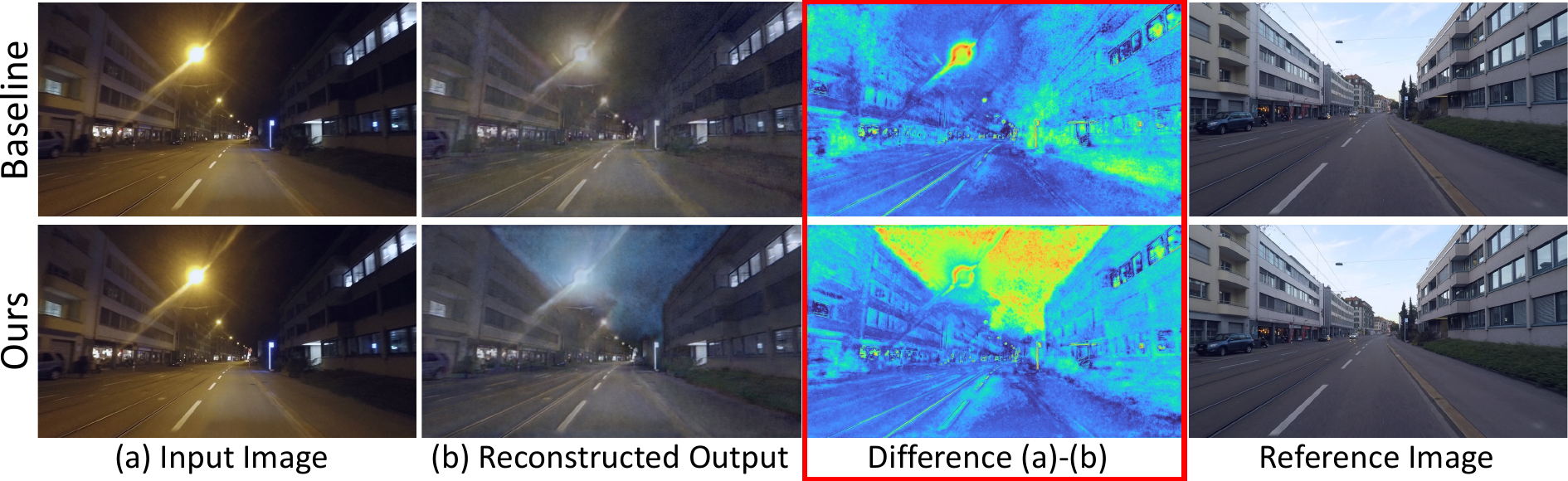}}
    \caption{Qualitative analysis on FREST.}
    \label{fig:hist}
\end{figure}

We present a detailed analysis on FREST.
Please note that the goal of FREST is not image restoration but feature restoration, which means FREST aims to train a model to robustly recognize adverse condition images as if they were in normal conditions, not to convert them into normal images.
As shown in Fig.~6 and \Fig{hist}, we reconstructed images only to qualitatively investigate the impact of feature restoration.
They show the favorable impact of FREST on recognition, particularly in enhancing the boundaries of buildings, compared to the baseline results.

\subsection{Analysis on Adverse Condition Discriminating Loss}
We devised the adverse condition discriminating loss to further mitigate the adverse effects of our segmentation feature, which is computed from the segmentation encoder.
To implement this strategy, we introduced a condition discriminator which classifies each condition (\ie, adverse and normal conditions).
For in-depth analysis, we introduce other possible solutions to remove the adverse effects.

\begin{figure*}[h]
    \centering
    \vspace{0mm}
    \scalebox{1}{
    \includegraphics[width=1\linewidth]{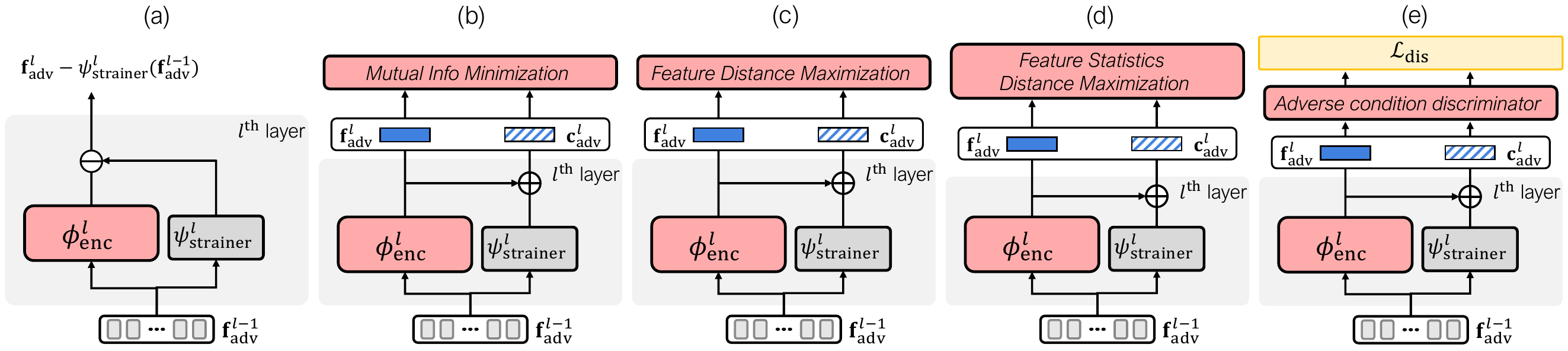} }
\caption{
Illustrations for the possible solutions of removing adverse effects from the segmentation feature computed by the segmentation encoder.
(a) Residual learning (b) Minimizing mutual information (c) Maximizing feature distance (d) Maximizing feature statistics distance (e) Feature classification (Ours)
} \label{fig:dis_loss}
\end{figure*}

Our initial approach is employing residual learning~\cite{he2016deep} to eliminate condition-specific information from the encoder features $\textbf{f}_\textrm{adv}$ as illustrated in \Fig{dis_loss}(a).
Additionally, we implement a strategy of minimizing mutual information~\cite{cheng2020club} between the encoder feature $\textbf{f}_\textrm{adv}$ and condition-specific feature $\textbf{c}_\textrm{adv}$ as shown in \Fig{dis_loss}(b).
Furthermore, we maximize the feature distance between the encoder feature $\textbf{f}_\textrm{adv}$ and condition-specific feature $\textbf{c}_\textrm{adv}$ by utilizing L1 loss as \Fig{dis_loss}(c).
Finally, to utilize the domain information more effectively, we maximize the feature statistics distance.
For this, we calculate the mean and standard deviation of features as feature statistics and maximize feature statistics between $\textbf{f}_\textrm{adv}$ and $\textbf{c}_\textrm{adv}$ as \Fig{dis_loss}(d).
Subsequently, through empirical analysis, we demonstrate empirically that our adverse condition discriminating loss, implemented through feature classification, is the most effective method when compared to the aforementioned possible solutions.

\begin{table}[h]
    \centering
    \renewcommand{\arraystretch}{1.0}    
    \caption{
    Analysis of possible solutions for adverse condition distancing loss. The results are reported in mIoU on the ACDC validation set.
    }
    \scalebox{1.0}{\begin{tabular}{lc}
    \toprule
    Possible Solution & mIoU \\
    \midrule
       (a) Residual learning        &  62.8      \\
       (b) Minimizing mutual information  &  65.6   \\
       (c) Maximizing feature distance   &  66.5   \\
       (d) Maximizing feature statistics distance  &   67.4   \\
       (e) Feature classification (Ours)   & \textbf{68.6}       \\
     \bottomrule
    \end{tabular}}
    \label{tab:abl_classifier}
\end{table}

\section{Additional Ablation Study}\label{appendix:ablation}

\subsection{Effect of Each Loss}
We investigate contributions of entropy loss $\mathcal{L}_\textrm{ent}$, self-training loss $\mathcal{L}_\textrm{self}$, and our proposed losses (\ie~feature restoration loss $\mathcal{L}_\textrm{resto}$ and adverse condition discriminating loss $\mathcal{L}_\textrm{dis}$) on ACDC~\cite{acdc} validation performance.
To this end, we evaluate our model without each loss.
As presented in \Tbl{loss}, all the losses contribute to the performance.
Especially, each impact of $\mathcal{L}_\textrm{self}$ and our proposed losses (\ie~feature restoration loss $\mathcal{L}_\textrm{resto}$ and adverse condition discriminating loss $\mathcal{L}_\textrm{dis}$) is larger than another component.

\begin{table}[h!]
    \centering
    \renewcommand{\arraystretch}{1.0}
    \caption{Effect of additional losses as self-training and entropy minimization loss. The results are reported in mIoU on the ACDC validation set.
    }
    \scalebox{1.0}{\begin{tabular}{cccc}
    \toprule
    w/o $\mathcal{L}_\textrm{self}$ & w/o $\mathcal{L}_\textrm{ent}$ & w/o $\mathcal{L}_\textrm{resto}\&\mathcal{L}_\textrm{dis}$ & mIoU \\ 
    \midrule
    \checkmark  &      &               &  59.1   \\
         & \checkmark  &               &  67.8   \\
         &             &   \checkmark  &  62.7  \\
         &             &               &  \textbf{68.6}  \\
     \bottomrule
    \end{tabular}}
    \label{tab:loss}
\end{table}

\subsection{Effect of Patch Confidence Threshold}
\Tbl{condition_specific} presents the sensitivity of FREST performance to the confidence threshold value in Eq.~(3) and Eq.~(4).
These results show that our method is insensitive to the hyperparameter of confidence.

\begin{table}[h!]
    \centering
    \renewcommand{\arraystretch}{1.0}
    \caption{Effect of the confidence threshold on FREST. The results are reported in mIoU on the ACDC validation set.}
    \scalebox{1.0}{\begin{tabular}{lccccccc}
    \toprule
    ~Confidence Threshold ~&~ 0 ~&~ 0.1 ~&~ 0.2 ~&~ 0.3 ~&~ 0.4 ~&~ 0.5 ~&~ 0.6~ \\ 
    \midrule
     ~mIoU ~&~ 67.5 ~&~  68.2 ~&~ \textbf{68.6} ~&~ 68.0 ~&~ 67.7 ~&~ 67.3 ~&~ 67.4~  \\
     \bottomrule
    \end{tabular}}
    \label{tab:condition_specific}
\end{table}

\section{Condition-Wise Performance}\label{appendix:conditionwise}
In this section, we present the condition-wise test results on ACDC Fog, ACDC Night, ACDC Rain, and ACDC Snow~\cite{acdc}.
In \Tbl{acdc_fog}, \ref{tab:acdc_night}, \ref{tab:acdc_rain}, \ref{tab:acdc_snow}, FREST outperforms all the competitors in the four condition splits.

\section{Application to a Different Backbone.}
We employed DeepLab-v2~\cite{deeplab_v2} as the segmentation backbone, and modified our condition strainer structure by substituting its linear layers with 1$\times$1 convolution layers for this ConvNet architecture.
As shown in \Tbl{backbone}, FREST significantly outperformed both the source model and CMA in this setting as well, suggesting that it is generic enough.

\begin{table}[h!]
  \centering
    \caption{Results of application to a different backbone. The results are reported in mIoU on the ACDC validation set.}
  \resizebox{0.4\linewidth}{!}{
  \begin{tabular}{@{}ccc@{}}
  \toprule
   Source model	& CMA	&FREST	\\
    \midrule
   37.6 & 46.6	& \textbf{48.4}  \\
  \bottomrule
  \end{tabular}
  }\label{tab:backbone}
\end{table}

\section{Impact of the Length of the Positive Queue.}
We conducted an ablation study to investigate the impact of the length of the positive queue.
As shown in \Tbl{queue_length}, FREST is insensitive to the length.
\begin{table}[h]
  \centering
    \caption{Effect of the length of the positive queue. The results are reported in mIoU on the ACDC validation set.}
  \resizebox{0.7\linewidth}{!}{
  \begin{tabular}{@{}lccccccc@{}}
  \toprule
  Length  &50K	&55K	&60K	& 65K (Ours)	&70K	&75K    &80K\\
    \midrule
    & 68.3 & 68.0	& 68.4  & 68.6 	& 68.1 & 68.2  & 67.9\\
  \bottomrule
  \end{tabular}}
  \label{tab:queue_length} 
\end{table}

\section{Additional Explanation for Fig.~7.}
\begin{figure}[h]
    \centering
    \scalebox{0.4}{
    \includegraphics[width=\linewidth]{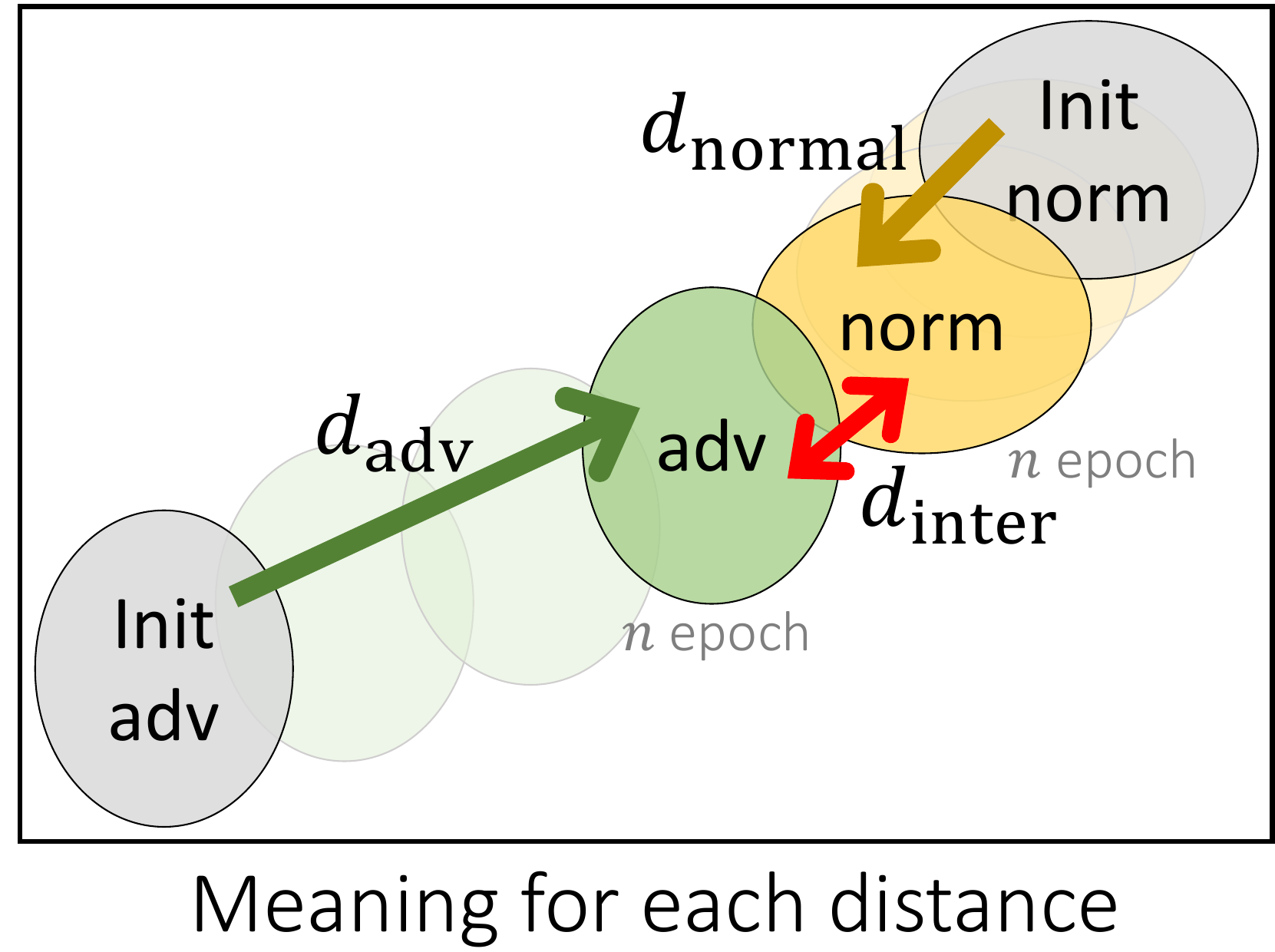}}
    \caption{Meaning of each distance in Fig.~7 of the main paper.}
    \label{fig:recon_supp}
\end{figure}

This section presents an additional explanation for the meaning of Fig.~7 in the main paper.
It is natural for the intra-domain distance to increase since, as a model updates, the feature distribution changes from the initial distribution during training.
Notably, as seen in Fig.~7 and \Fig{recon_supp}, we included for clarity, \greend{$d_{\text{adv}}$} increases significantly more than \yellowd{$d_{\text{normal}}$}, while \redd{$d_{\text{inter}}$} between normal and adverse conditions decreases.
This suggests the desired feature restoration: adverse condition features shift towards normal condition features during training with FREST.

\section{Computational resource and training time.}
FREST was trained using a single NVIDIA RTX 3090 GPU, taking 9 hours and 45 minutes, significantly faster than UDA methods~\cite{AdaptSegNet,vu2019advent} which typically take about 4 days. 
Our method is particularly efficient during inference as it uses only the segmentation backbone without any auxiliary modules.

\section{Additional Qualitative Results}\label{appendix:additional_qual}
\begin{figure*}[t]
    \centering
    \vspace{0mm}
    \scalebox{1}{
    \includegraphics[width=0.98\linewidth]{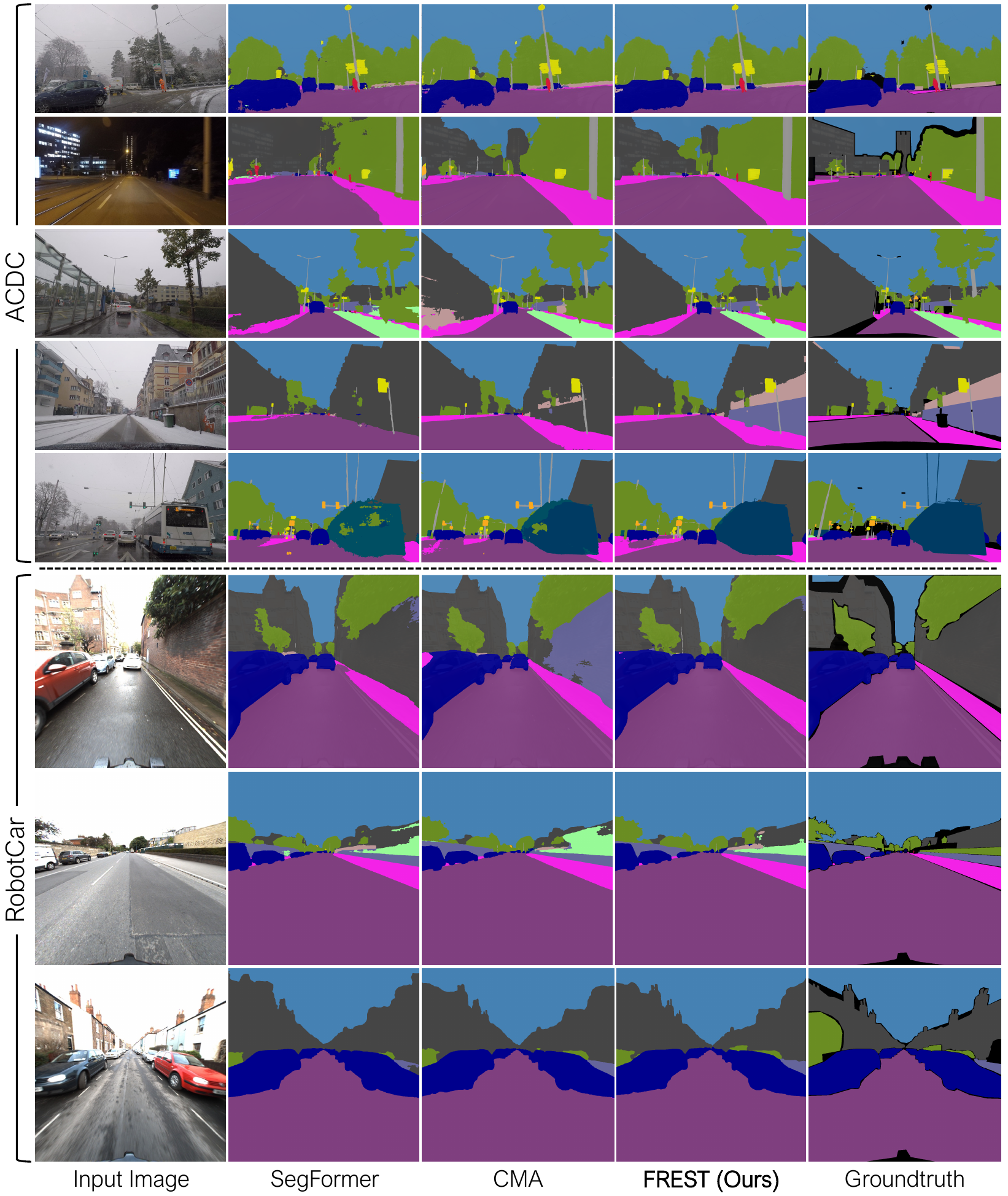} }
\caption{
Qualitative segmentation results for ACDC and RobotCar.
} \label{fig:qual}
\end{figure*}

We present more qualitative results on ACDC~\cite{acdc} and Robotcar~\cite{larsson2019cross,maddern20171} in this section.
\Fig{qual} shows the results of SegFormer, CMA, and FREST (Ours).
This demonstrates that SegFormer and CMA often fail to predict detailed objects, while our method surpasses them.

\clearpage

\begin{table*}[t!]
\smallskip
\centering
\caption{
Comparison with source-free DA methods on Cityscapes$\to$ACDC.
The results are reported in mIoU (\%) on the ACDC Fog test set.
}
\scalebox{0.63}{
\ra{1.13}
\begin{tabular}{@{}lcccccccccccccccccccr@{}}\toprule
\multirow{3}{*}{Method} & \multicolumn{19}{c}{ACDC Fog IoU} \\
\cmidrule{2-21} & \rotatebox[origin=c]{90}{road} & \rotatebox[origin=c]{90}{sidew.} & \rotatebox[origin=c]{90}{build.} & \rotatebox[origin=c]{90}{wall} & \rotatebox[origin=c]{90}{fence} & \rotatebox[origin=c]{90}{pole} & \rotatebox[origin=c]{90}{light} & \rotatebox[origin=c]{90}{sign} & \rotatebox[origin=c]{90}{veget.} & \rotatebox[origin=c]{90}{terrain} & \rotatebox[origin=c]{90}{sky} & \rotatebox[origin=c]{90}{person} & \rotatebox[origin=c]{90}{rider} & \rotatebox[origin=c]{90}{car} & \rotatebox[origin=c]{90}{truck} & \rotatebox[origin=c]{90}{bus} & \rotatebox[origin=c]{90}{train} & \rotatebox[origin=c]{90}{motorc.} & \rotatebox[origin=c]{90}{bicycle} & \multicolumn{1}{c}{\phantom{00}\rotatebox[origin=c]{90}{\textbf{mean}}} \\ \midrule
Source model~\cite{xie2021segformer}  & 87.8 & 60.7 & 73.1 & 44.5 & 30.1 & 42.1 & 52.3 & 64.4 & 81.4 & 68.8 & 93.4 & 51.1 & 53.2 & 78.4 & 66.0 & 39.7 & 75.1 & 43.2 & 47.4 & 60.7 \\
HCL~\cite{huang2021model} & 88.5 & 63.2 & 79.8 & 45.3 & 30.6 & 44.7 & 53.7 & 65.9 & 81.8 & 69.6 & 95.5 & 52.5 & 55.0 & 79.4 & 68.0 & 40.7 & 74.0 & 40.7 & 46.9 & 61.9 \\
URMA~\cite{fleuret2021uncertainty} &  89.3 & 61.8 & \underline{87.9} & 51.4 & \underline{36.3} & \underline{52.3} & 58.1 & \underline{67.9} & 85.7 & \underline{71.8} & 97.2 & 54.5 & \textbf{62.5} & \underline{82.3} & \underline{70.6} & \underline{62.0} & 82.0 & \underline{52.9} & 36.2 & \underline{66.5} \\
CMA~\cite{bruggemann2023contrastive} & \textbf{93.5} & \textbf{75.3} & \textbf{88.6} & \textbf{53.4} & 33.0 & 52.2 & \underline{58.2} & 67.0 & \underline{86.9} & 71.5 & \textbf{97.8} & \underline{55.6} & 42.0 & 80.4 & 70.0 & 54.8 & \underline{83.3} & 43.0 & \underline{37.4} & 65.5 \\
FREST & \textbf{93.5} & \underline{74.4} & 87.4 & \underline{51.5} & \textbf{36.7} & \textbf{54.1} & \textbf{59.1} & \textbf{69.6} & \textbf{87.2} & \textbf{72.1} & \underline{97.6} & \textbf{59.8} & \underline{60.1} & \textbf{85.1} & \textbf{73.8} & \textbf{77.2} & \textbf{84.7} & \textbf{63.6} & \textbf{45.6}  & \textbf{70.2} \\
\bottomrule
\end{tabular}
}
\label{tab:acdc_fog}
\end{table*}

\begin{table*}[t!]
\smallskip
\centering
\caption{
Comparison with source-free DA methods on Cityscapes$\to$ACDC.
The results are reported in mIoU (\%) on the ACDC Night test set.
}
\scalebox{0.63}{
\ra{1.13}
\begin{tabular}{@{}lcccccccccccccccccccr@{}}\toprule
\multirow{3}{*}{Method} & \multicolumn{19}{c}{ACDC Night IoU} \\
\cmidrule{2-21} & \rotatebox[origin=c]{90}{road} & \rotatebox[origin=c]{90}{sidew.} & \rotatebox[origin=c]{90}{build.} & \rotatebox[origin=c]{90}{wall} & \rotatebox[origin=c]{90}{fence} & \rotatebox[origin=c]{90}{pole} & \rotatebox[origin=c]{90}{light} & \rotatebox[origin=c]{90}{sign} & \rotatebox[origin=c]{90}{veget.} & \rotatebox[origin=c]{90}{terrain} & \rotatebox[origin=c]{90}{sky} & \rotatebox[origin=c]{90}{person} & \rotatebox[origin=c]{90}{rider} & \rotatebox[origin=c]{90}{car} & \rotatebox[origin=c]{90}{truck} & \rotatebox[origin=c]{90}{bus} & \rotatebox[origin=c]{90}{train} & \rotatebox[origin=c]{90}{motorc.} & \rotatebox[origin=c]{90}{bicycle} & \multicolumn{1}{c}{\phantom{00}\rotatebox[origin=c]{90}{\textbf{mean}}} \\ \midrule
Source model~\cite{xie2021segformer}&87.9 & 52.7 & 64.1 & 34.0 & 20.2 & 37.2 & 34.5 & 40.2 & 51.8 & 32.4 & 6.6 & 54.5 & 31.4 & 72.8 & 49.6 & 65.2 & 54.1 & 34.0 & 41.4 & 45.5\\
HCL~\cite{huang2021model}& 88.2 & 54.3 & 64.4 & 35.3 & 20.7 & 39.1 & 36.8 & 40.4 & 52.0 & 32.1 & 2.8 & 55.2 & 33.7 & 73.5 & 49.2 & 66.5 & 58.1 & 35.4 & 41.7 & 46.3\\
URMA~\cite{fleuret2021uncertainty}& 90.6 & 60.1 & 71.9 & 42.6 & 26.7 & 47.5 & 47.5 & 47.4 & 46.7 & 42.9 & 0.4 & 54.4 & 34.6 & 76.8 & 42.1 & 65.6 & 71.0 & \underline{38.0} & 37.2 & 49.7\\
CMA~\cite{bruggemann2023contrastive}&\textbf{95.2} & \textbf{77.5} & \textbf{84.3} & \underline{43.9} & \underline{30.9} & \underline{49.4} & \textbf{52.0} & \textbf{49.6} & \textbf{74.2} & \textbf{51.2} & \underline{78.4} & \underline{61.4} & \underline{41.2} & \underline{79.2} & \underline{63.6} & \underline{75.1} & \textbf{75.8} & 34.6 & \underline{47.3} & \underline{61.3}\\
FREST & \underline{94.6} & \underline{75.1} & \underline{82.5} & \textbf{44.2} & \textbf{32.8} & \textbf{53.2} & \underline{48.5} & \underline{49.2} & \underline{71.1} & \underline{48.5} & \textbf{78.5} & \textbf{63.0} & \textbf{41.5} & \textbf{82.7} & \textbf{67.1} & \textbf{75.5} & \underline{74.6} & \textbf{48.3} & \textbf{50.7}  & \textbf{62.2} \\
\bottomrule
\end{tabular}
}
\label{tab:acdc_night}
\end{table*}

\begin{table*}[t!]
\smallskip
\centering
\caption{
Comparison with source-free DA methods on Cityscapes$\to$ACDC.
The results are reported in mIoU (\%) on the ACDC Rain test set.
}
\scalebox{0.63}{
\ra{1.13}
\begin{tabular}{@{}lcccccccccccccccccccr@{}}\toprule
\multirow{3}{*}{Method} & \multicolumn{19}{c}{ACDC Rain IoU} \\
\cmidrule{2-21} & \rotatebox[origin=c]{90}{road} & \rotatebox[origin=c]{90}{sidew.} & \rotatebox[origin=c]{90}{build.} & \rotatebox[origin=c]{90}{wall} & \rotatebox[origin=c]{90}{fence} & \rotatebox[origin=c]{90}{pole} & \rotatebox[origin=c]{90}{light} & \rotatebox[origin=c]{90}{sign} & \rotatebox[origin=c]{90}{veget.} & \rotatebox[origin=c]{90}{terrain} & \rotatebox[origin=c]{90}{sky} & \rotatebox[origin=c]{90}{person} & \rotatebox[origin=c]{90}{rider} & \rotatebox[origin=c]{90}{car} & \rotatebox[origin=c]{90}{truck} & \rotatebox[origin=c]{90}{bus} & \rotatebox[origin=c]{90}{train} & \rotatebox[origin=c]{90}{motorc.} & \rotatebox[origin=c]{90}{bicycle} & \multicolumn{1}{c}{\phantom{00}\rotatebox[origin=c]{90}{\textbf{mean}}} \\ \midrule
Source model~\cite{xie2021segformer}&83.1 & 46.7 & 89.5 & 40.5 & 47.2 & 54.0 & 67.0 & 66.9 & 92.6 & 40.2 & 97.6 & 63.5 & 24.6 & 87.8 & 65.1 & 72.7 & 81.0 & 42.8 & 58.0 & 64.3\\
HCL~\cite{huang2021model}&84.2 & 50.5 & 90.1 & 42.7 & 48.9 & 57.0 & 68.5 & 69.0 & 93.0 & 40.9 & 97.8 & 65.4 & 26.1 & 88.7 & 68.1 & 74.4 & 80.4 & 43.8 & 58.0 & 65.6\\
URMA~\cite{fleuret2021uncertainty}&87.2 & 61.0 & 92.4 & 52.0 & 51.9 & 57.2 & \underline{72.0} & \underline{73.1} & \underline{93.8} & \textbf{46.1} & \underline{98.1} & \underline{68.8} & 31.8 & \underline{90.6} & \underline{73.2} & \underline{85.9} & \textbf{86.9} & \underline{51.7} & 51.9 & 69.8\\
CMA~\cite{bruggemann2023contrastive}&\textbf{93.3} & \textbf{76.3} & \underline{92.8} & \underline{58.1} & \textbf{58.2} & \underline{61.2} & 70.4 & 71.8 & \underline{93.8} & \underline{45.0} & 97.9 & 67.4 & \underline{36.8} & 89.7 & 72.2 & \textbf{88.5} & \underline{86.4} & 50.5 & \textbf{66.7} & \underline{72.5}\\
FREST &\underline{92.1} & \underline{73.5} & \textbf{93.9} & \textbf{62.3} & \underline{57.8} & \textbf{65.4} & \textbf{72.7} & \textbf{75.8} & \textbf{93.9} & 42.2 & \textbf{98.4} & \textbf{72.4} & \textbf{39.0} & \textbf{92.6} & \textbf{79.4} & 84.5 & 84.9 & \textbf{55.6} & \underline{65.4} & \textbf{73.8} \\
\bottomrule
\end{tabular}
}
\label{tab:acdc_rain}
\end{table*}

\begin{table*}[t!]
\smallskip
\centering
\caption{
Comparison with source-free DA methods on Cityscapes$\to$ACDC.
The results are reported in mIoU (\%) on the ACDC Snow test set.
}
\scalebox{0.63}{
\ra{1.13}
\begin{tabular}{@{}lcccccccccccccccccccr@{}}\toprule
\multirow{3}{*}{Method} & \multicolumn{19}{c}{ACDC Snow IoU} \\
\cmidrule{2-21} & \rotatebox[origin=c]{90}{road} & \rotatebox[origin=c]{90}{sidew.} & \rotatebox[origin=c]{90}{build.} & \rotatebox[origin=c]{90}{wall} & \rotatebox[origin=c]{90}{fence} & \rotatebox[origin=c]{90}{pole} & \rotatebox[origin=c]{90}{light} & \rotatebox[origin=c]{90}{sign} & \rotatebox[origin=c]{90}{veget.} & \rotatebox[origin=c]{90}{terrain} & \rotatebox[origin=c]{90}{sky} & \rotatebox[origin=c]{90}{person} & \rotatebox[origin=c]{90}{rider} & \rotatebox[origin=c]{90}{car} & \rotatebox[origin=c]{90}{truck} & \rotatebox[origin=c]{90}{bus} & \rotatebox[origin=c]{90}{train} & \rotatebox[origin=c]{90}{motorc.} & \rotatebox[origin=c]{90}{bicycle} & \multicolumn{1}{c}{\phantom{00}\rotatebox[origin=c]{90}{\textbf{mean}}} \\ \midrule
Source model~\cite{xie2021segformer}&82.0 & 44.9 & 80.5 & 30.4 & 45.4 & 46.8 & 65.6 & 63.1 & 86.8 & 5.2 & 93.6 & 67.8 & 40.8 & 87.1 & 56.4 & 76.7 & 83.1 & 32.8 & 60.3 & 60.5\\
HCL~\cite{huang2021model}&82.9 & 47.4 & 83.2 & 35.4 & 46.8 & 50.1 & 67.8 & 64.9 & 87.7 & 5.3 & 95.6 & 69.8 & 43.9 & 87.6 & 60.1 & 76.9 & 83.2 & 35.3 & 63.4 & 62.5\\
URMA~\cite{fleuret2021uncertainty}&88.0 & 58.9 & 87.2 & \underline{52.0} & 51.7 & \underline{57.8} & \underline{75.6} & 70.3 & 88.8 & 5.8 & \underline{97.1} & 75.0 & \underline{63.6} & 89.0 & \underline{69.6} & 79.0 & \textbf{89.8} & \underline{50.1} & 65.4 & 69.2\\
CMA~\cite{bruggemann2023contrastive}&\textbf{92.4} & \textbf{70.5} & \underline{88.3} & 50.4 & \textbf{55.6} & 56.3 & 74.8 & \underline{71.1} & \textbf{90.8} & \textbf{29.4} & 96.9 & \underline{77.4} & 63.5 & \underline{90.1} & 63.5 & \underline{79.6} & \underline{89.0} & 45.6 & \textbf{73.9} & \underline{71.5}\\
FREST&\underline{91.3} & \underline{65.0} & \textbf{88.4} & \textbf{54.5} & \underline{55.3} & \textbf{60.8} & \textbf{76.6} & \textbf{73.9} & \underline{89.6} & \underline{10.6} & \textbf{97.4} & \textbf{79.6} & \textbf{66.3} & \textbf{91.8} & \textbf{72.4} & \textbf{80.4} & 88.3 & \textbf{53.6} & \underline{72.8} & \textbf{72.0}\\
\bottomrule
\end{tabular}
}
\label{tab:acdc_snow}
\end{table*}

\clearpage  

{\small
\bibliographystyle{splncs04}
\bibliography{paper}

\begin{thebibliography}{10}
\providecommand{\url}[1]{\texttt{#1}}
\providecommand{\urlprefix}{URL }
\providecommand{\doi}[1]{https://doi.org/#1}

\bibitem{bruggemann2023contrastive}
Br{\"u}ggemann, D., Sakaridis, C., Br{\"o}dermann, T., Van~Gool, L.: Contrastive model adaptation for cross-condition robustness in semantic segmentation. In: Proc. IEEE/CVF International Conference on Computer Vision (ICCV) (2023)

\bibitem{bruggemann2023refign}
Br{\"u}ggemann, D., Sakaridis, C., Truong, P., Van~Gool, L.: Refign: Align and refine for adaptation of semantic segmentation to adverse conditions. In: Proc. IEEE Winter Conference on Applications of Computer Vision (WACV) (2023)

\bibitem{deeplab_v2}
Chen, L.C., Papandreou, G., Kokkinos, I., Murphy, K., Yuille, A.L.: Deeplab: Semantic image segmentation with deep convolutional nets, atrous convolution, and fully connected crfs. IEEE Transactions on Pattern Analysis and Machine Intelligence (TPAMI)  (2017)

\bibitem{adaptformer}
Chen, S., Ge, C., Tong, Z., Wang, J., Song, Y., Wang, J., Luo, P.: Adaptformer: Adapting vision transformers for scalable visual recognition. In: Proc. Neural Information Processing Systems (NeurIPS) (2022)

\bibitem{chen2018cascaded}
Chen, Y., Wang, Z., Peng, Y., Zhang, Z., Yu, G., Sun, J.: Cascaded pyramid network for multi-person pose estimation. In: Proc. IEEE/CVF Conference on Computer Vision and Pattern Recognition (CVPR) (2018)

\bibitem{cheng2020club}
Cheng, P., Hao, W., Dai, S., Liu, J., Gan, Z., Carin, L.: Club: A contrastive log-ratio upper bound of mutual information. In: Proc. International Conference on Machine Learning (ICML) (2020)

\bibitem{choi2021robustnet}
Choi, S., Jung, S., Yun, H., Kim, J.T., Kim, S., Choo, J.: Robustnet: Improving domain generalization in urban-scene segmentation via instance selective whitening. In: Proc. IEEE/CVF Conference on Computer Vision and Pattern Recognition (CVPR) (2021)

\bibitem{cityscapes}
Cordts, M., Omran, M., Ramos, S., Rehfeld, T., Enzweiler, M., Benenson, R., Franke, U., Roth, S., Schiele, B.: The cityscapes dataset for semantic urban scene understanding. In: Proc. IEEE/CVF Conference on Computer Vision and Pattern Recognition (CVPR) (2016)

\bibitem{dai2020curriculum}
Dai, D., Sakaridis, C., Hecker, S., Van~Gool, L.: Curriculum model adaptation with synthetic and real data for semantic foggy scene understanding. International Journal of Computer Vision (IJCV)  (2020)

\bibitem{dosovitskiy2020image}
Dosovitskiy, A., Beyer, L., Kolesnikov, A., Weissenborn, D., Zhai, X., Unterthiner, T., Dehghani, M., Minderer, M., Heigold, G., Gelly, S., et~al.: An image is worth 16x16 words: Transformers for image recognition at scale. In: Proc. International Conference on Learning Representations (ICLR) (2021)

\bibitem{guo2022simt}
Guo, X., Liu, J., Liu, T., Yuan, Y.: Simt: Handling open-set noise for domain adaptive semantic segmentation. In: Proc. IEEE/CVF Conference on Computer Vision and Pattern Recognition (CVPR) (2022)

\bibitem{he2016deep}
He, K., Zhang, X., Ren, S., Sun, J.: Deep residual learning for image recognition. In: Proc. IEEE/CVF Conference on Computer Vision and Pattern Recognition (CVPR) (2016)

\bibitem{hoffman2016fcns}
Hoffman, J., Wang, D., Yu, F., Darrell, T.: Fcns in the wild: Pixel-level adversarial and constraint-based adaptation. arXiv preprint arXiv:1612.02649  (2016)

\bibitem{houlsby2019parameter}
Houlsby, N., Giurgiu, A., Jastrzebski, S., Morrone, B., De~Laroussilhe, Q., Gesmundo, A., Attariyan, M., Gelly, S.: Parameter-efficient transfer learning for nlp. In: Proc. International Conference on Machine Learning (ICML). PMLR (2019)

\bibitem{hoyer2022daformer}
Hoyer, L., Dai, D., Van~Gool, L.: {DAFormer}: Improving network architectures and training strategies for domain-adaptive semantic segmentation. In: Proc. IEEE/CVF Conference on Computer Vision and Pattern Recognition (CVPR) (2022)

\bibitem{hoyer2022hrda}
Hoyer, L., Dai, D., Van~Gool, L.: {HRDA}: Context-aware high-resolution domain-adaptive semantic segmentation. In: Proc. European Conference on Computer Vision (ECCV) (2022)

\bibitem{hoyer2023mic}
Hoyer, L., Dai, D., Wang, H., Van~Gool, L.: Mic: Masked image consistency for context-enhanced domain adaptation. In: Proc. IEEE/CVF Conference on Computer Vision and Pattern Recognition (CVPR). pp. 11721--11732 (2023)

\bibitem{huang2021model}
Huang, J., Guan, D., Xiao, A., Lu, S.: Model adaptation: Historical contrastive learning for unsupervised domain adaptation without source data. In: Proc. Neural Information Processing Systems (NeurIPS) (2021)

\bibitem{huttenlocher1993comparing}
Huttenlocher, D.P., Klanderman, G.A., Rucklidge, W.J.: Comparing images using the hausdorff distance. IEEE Transactions on Pattern Analysis and Machine Intelligence (TPAMI)  (1993)

\bibitem{khosla2020supervised}
Khosla, P., Teterwak, P., Wang, C., Sarna, A., Tian, Y., Isola, P., Maschinot, A., Liu, C., Krishnan, D.: Supervised contrastive learning. In: Proc. Neural Information Processing Systems (NeurIPS) (2020)

\bibitem{kim2020learning}
Kim, M., Byun, H.: Learning texture invariant representation for domain adaptation of semantic segmentation. In: Proc. IEEE/CVF Conference on Computer Vision and Pattern Recognition (CVPR) (2020)

\bibitem{larsson2019cross}
Larsson, M., Stenborg, E., Hammarstrand, L., Pollefeys, M., Sattler, T., Kahl, F.: A cross-season correspondence dataset for robust semantic segmentation. In: Proc. IEEE/CVF Conference on Computer Vision and Pattern Recognition (CVPR) (2019)

\bibitem{lee2023human}
Lee, S., Rim, J., Jeong, B., Kim, G., Woo, B., Lee, H., Cho, S., Kwak, S.: Human pose estimation in extremely low-light conditions. In: Proc. IEEE/CVF Conference on Computer Vision and Pattern Recognition (CVPR) (2023)

\bibitem{lee2022fifo}
Lee, S., Son, T., Kwak, S.: Fifo: Learning fog-invariant features for foggy scene segmentation. In: Proc. IEEE/CVF Conference on Computer Vision and Pattern Recognition (CVPR) (2022)

\bibitem{li2019bidirectional}
Li, Y., Yuan, L., Vasconcelos, N.: Bidirectional learning for domain adaptation of semantic segmentation. In: Proc. IEEE/CVF Conference on Computer Vision and Pattern Recognition (CVPR) (2019)

\bibitem{li2018megadepth}
Li, Z., Snavely, N.: Megadepth: Learning single-view depth prediction from internet photos. In: Proc. IEEE/CVF Conference on Computer Vision and Pattern Recognition (CVPR) (2018)

\bibitem{liu2021source}
Liu, Y., Zhang, W., Wang, J.: Source-free domain adaptation for semantic segmentation. In: Proc. IEEE/CVF Conference on Computer Vision and Pattern Recognition (CVPR) (2021)

\bibitem{loshchilov2017decoupled}
Loshchilov, I., Hutter, F.: Decoupled weight decay regularization. In: Proc. International Conference on Learning Representations (ICLR) (2019)

\bibitem{van2008visualizing}
Van~der Maaten, L., Hinton, G.: Visualizing data using t-sne. Journal of machine learning research  (2008)

\bibitem{maddern20171}
Maddern, W., Pascoe, G., Linegar, C., Newman, P.: 1 year, 1000 km: The oxford robotcar dataset. The International Journal of Robotics Research  (2017)

\bibitem{pan2020unsupervised}
Pan, F., Shin, I., Rameau, F., Lee, S., Kweon, I.S.: Unsupervised intra-domain adaptation for semantic segmentation through self-supervision. In: Proc. IEEE/CVF Conference on Computer Vision and Pattern Recognition (CVPR) (2020)

\bibitem{pfeiffer2020adapterfusion}
Pfeiffer, J., Kamath, A., R{\"u}ckl{\'e}, A., Cho, K., Gurevych, I.: Adapterfusion: Non-destructive task composition for transfer learning. arXiv preprint arXiv:2005.00247  (2020)

\bibitem{gta5}
Richter, S.R., Vineet, V., Roth, S., Koltun, V.: Playing for data: Ground truth from computer games. In: Proc. European Conference on Computer Vision (ECCV) (2016)

\bibitem{synthia}
Ros, G., Sellart, L., Materzynska, J., Vazquez, D., Lopez, A.M.: The synthia dataset: A large collection of synthetic images for semantic segmentation of urban scenes. In: Proc. IEEE/CVF Conference on Computer Vision and Pattern Recognition (CVPR) (2016)

\bibitem{Sakaridis_2019_ICCV}
Sakaridis, C., Dai, D., Gool, L.V.: Guided curriculum model adaptation and uncertainty-aware evaluation for semantic nighttime image segmentation. In: Proc. IEEE/CVF International Conference on Computer Vision (ICCV) (2019)

\bibitem{Sakaridis_2018_ECCV}
Sakaridis, C., Dai, D., Hecker, S., Van~Gool, L.: Model adaptation with synthetic and real data for semantic dense foggy scene understanding. In: Proc. European Conference on Computer Vision (ECCV) (2018)

\bibitem{Sakaridis_2018_IJCV}
Sakaridis, C., Dai, D., Van~Gool, L.: Semantic foggy scene understanding with synthetic data. International Journal of Computer Vision (IJCV)  (2018)

\bibitem{acdc}
Sakaridis, C., Dai, D., {Van Gool}, L.: {ACDC}: The adverse conditions dataset with correspondences for semantic driving scene understanding. In: Proc. IEEE/CVF International Conference on Computer Vision (ICCV) (2021)

\bibitem{son2020urie}
Son, T., Kang, J., Kim, N., Cho, S., Kwak, S.: Urie: Universal image enhancement for visual recognition in the wild. In: Proc. European Conference on Computer Vision (ECCV) (2020)

\bibitem{sun2019deep}
Sun, K., Xiao, B., Liu, D., Wang, J.: Deep high-resolution representation learning for human pose estimation. In: Proc. IEEE/CVF Conference on Computer Vision and Pattern Recognition (CVPR) (2019)

\bibitem{fleuret2021uncertainty}
Teja~S, P., Fleuret, F.: Uncertainty reduction for model adaptation in semantic segmentation. In: Proc. IEEE/CVF Conference on Computer Vision and Pattern Recognition (CVPR) (2021)

\bibitem{toshev2014deeppose}
Toshev, A., Szegedy, C.: Deeppose: Human pose estimation via deep neural networks. In: Proc. IEEE/CVF Conference on Computer Vision and Pattern Recognition (CVPR) (2014)

\bibitem{tsai2018learning}
Tsai, Y.H., Hung, W.C., Schulter, S., Sohn, K., Yang, M.H., Chandraker, M.: Learning to adapt structured output space for semantic segmentation. In: Proc. IEEE/CVF Conference on Computer Vision and Pattern Recognition (CVPR) (2018)

\bibitem{AdaptSegNet}
Tsai, Y.H., Hung, W.C., Schulter, S., Sohn, K., Yang, M.H., Chandraker, M.: Learning to adapt structured output space for semantic segmentation. In: Proc. IEEE/CVF Conference on Computer Vision and Pattern Recognition (CVPR) (2018)

\bibitem{tsai2019domain}
Tsai, Y.H., Sohn, K., Schulter, S., Chandraker, M.: Domain adaptation for structured output via discriminative patch representations. In: Proc. IEEE/CVF International Conference on Computer Vision (ICCV) (2019)

\bibitem{vaswani2017attention}
Vaswani, A., Shazeer, N., Parmar, N., Uszkoreit, J., Jones, L., Gomez, A.N., Kaiser, {\L}., Polosukhin, I.: Attention is all you need. In: Proc. Neural Information Processing Systems (NeurIPS) (2017)

\bibitem{vu2019advent}
Vu, T.H., Jain, H., Bucher, M., Cord, M., P{\'e}rez, P.: Advent: Adversarial entropy minimization for domain adaptation in semantic segmentation. In: Proc. IEEE/CVF Conference on Computer Vision and Pattern Recognition (CVPR) (2019)

\bibitem{wang2021tent}
Wang, D., Shelhamer, E., Liu, S., Olshausen, B., Darrell, T.: Tent: Fully test-time adaptation by entropy minimization. In: Proc. International Conference on Learning Representations (ICLR) (2021)

\bibitem{xiao2018simple}
Xiao, B., Wu, H., Wei, Y.: Simple baselines for human pose estimation and tracking. In: Proc. European Conference on Computer Vision (ECCV) (2018)

\bibitem{xie2021segformer}
Xie, E., Wang, W., Yu, Z., Anandkumar, A., Alvarez, J.M., Luo, P.: {SegFormer}: Simple and efficient design for semantic segmentation with transformers. In: Proc. Neural Information Processing Systems (NeurIPS) (2021)

\bibitem{Zendel_2018_ECCV}
Zendel, O., Honauer, K., Murschitz, M., Steininger, D., Fernandez~Dominguez, G.: Wilddash - creating hazard-aware benchmarks. In: Proc. European Conference on Computer Vision (ECCV) (2018)

\bibitem{zhang2019fast}
Zhang, F., Zhu, X., Ye, M.: Fast human pose estimation. In: Proc. IEEE/CVF Conference on Computer Vision and Pattern Recognition (CVPR) (2019)

\bibitem{zhang2017curriculum}
Zhang, Y., David, P., Gong, B.: Curriculum domain adaptation for semantic segmentation of urban scenes. In: Proc. IEEE/CVF International Conference on Computer Vision (ICCV) (2017)

\bibitem{zhao2023towards}
Zhao, D., Wang, S., Zang, Q., Quan, D., Ye, X., Jiao, L.: Towards better stability and adaptability: Improve online self-training for model adaptation in semantic segmentation. In: Proc. IEEE/CVF Conference on Computer Vision and Pattern Recognition (CVPR) (2023)

\bibitem{zou2018unsupervised}
Zou, Y., Yu, Z., Kumar, B., Wang, J.: Unsupervised domain adaptation for semantic segmentation via class-balanced self-training. In: Proc. European Conference on Computer Vision (ECCV) (2018)

\bibitem{zou2019confidence}
Zou, Y., Yu, Z., Liu, X., Kumar, B., Wang, J.: Confidence regularized self-training. In: Proc. IEEE/CVF International Conference on Computer Vision (ICCV) (2019)

\end{thebibliography}
}

\end{document}


\pagestyle{headings}
\mainmatter
\def\ECCVSubNumber{2065}  

\title{Human Pose Estimation in the Dark 
\it  ---Supplementary Material--- }

\titlerunning{ECCV-22 submission ID \ECCVSubNumber} 
\authorrunning{ECCV-22 submission ID \ECCVSubNumber} 
\author{Anonymous ECCV submission}
\institute{Paper ID \ECCVSubNumber}

\maketitle

\begin{figure*}[t]
    \centering
    \vspace{-1mm}
    \scalebox{0.95}{
    \includegraphics[width=\linewidth]{Supplementary/Figures/geometric_alignment.pdf} }
\caption{
(a) \& (b) Stereo-anaglyph images before and after geometric alignment. (c) \& (d) Magnified views of (a) and (b).
} \label{fig:geometric_alignment}
\vspace{0mm}
\end{figure*}

\section{Geometric Alignment of Our Camera System}

As described in our main manuscript, there may be a small amount of geometric misalignment between two camera modules in our system.
Moreover, while moving around the camera system collecting the dataset, the movement of the camera system may introduce additional geometric misalignment.
To resolve this, we captured a reference image pair of a static scene before collecting data every time we move the camera system.
Then, we estimated a homography matrix between them using Evangelidis \etal's method~\cite{Evangelidis-TPAMI08} and aligned the collected well-lit images using the estimated homography matrix.
\Fig{geometric_alignment}(c)-(d) visualize the effect of geometric alignment using stereo-anaglyph images where a pair of well-lit and scaled low-light images from the camera modules are visualized in red and cyan.
As the figure shows, even before the geometric alignment, images from our camera system have only a small amount of misalignment. Nevertheless, the geometric alignment can successfully resolve the remaining misalignment.


\section{Additionally Using Unlabeled Images in the PID Dataset}



While the experiments in the main manuscript are conducted with only human-labeled images in the PID dataset, 
the dataset also provides unlabeled images.
In this section, we empirically show that the unlabeled images in the PID dataset can be utilized to further improve the pose estimation performance.
To this end, we fine-tune the models of pose estimation~\cite{Xiao_2018_ECCV} and human detection~\cite{chen2018cascaded} methods using human-labeled data in the PID dataset. Then, we generate the pseudo-labels for the unlabeled images using the fine-tuned models, and construct a training set including both human-labeled and pseudo-labeled images.
Table~\ref{tab:dataset_pseudo_labeled_statistics} reports statistics of four splits of the resulting training set.
We then train our method using the training set and evaluate its performance.
As shown in Table~\ref{tab:additional_PID}, utilizing pseudo-labeled images further improves the pose estimation accuracy for all the cases.



\begin{table}[t]
\centering
\caption{
Statistics of the training set with both human-labeled and pseudo-labeled images.}
\vspace{1mm}
\renewcommand{\arraystretch}{1.0}
\scalebox{1}{
\begin{tabular}{lccc}
\toprule
Splits    & $\#$Videos & $\#$Frames & $\#$Instances \\ \midrule
LL-easy   & 328        & 6,143      & 45,702        \\
LL-normal & 342        & 6,367      & 40,764        \\
LL-hard   & 284        & 5,124      & 29,745        \\
WL        & 954        & 17,634     & 116,211       \\
\bottomrule
\end{tabular}
}
\vspace{0mm}
\label{tab:dataset_pseudo_labeled_statistics}
\end{table}

\begin{table}[t]
\centering
\vspace{-5mm}
\caption{
Quantitative results using additional pseudo-labeled images in the PID dataset.}
\vspace{1mm}
\renewcommand{\arraystretch}{1.0}
\scalebox{1}{\begin{tabular}{lcccc}
\toprule
AP@0.5:0.95     & LL-easy         &  LL-normal       &  LL-hard     &  WL        \\ \midrule
Human-labeled PID       & 29.7   & 18.0  & 6.7       &  69.4      \\
+ Additional pseudo-labeled PID       & 32.2 & 	18.2	  & 6.9	      &  77.0       \\ \bottomrule
\end{tabular}}
\vspace{0mm}
\label{tab:additional_PID}
\end{table}

\section{Empirical Analysis on Lighting Conditions}
Our method is based on the assumption that the neural style encodes the lighting condition.
In this section, we empirically verify this assumption by visualizing the distributions of Gram matrices computed from low-light and well-lit images.
\Fig{tsne} shows $t$-SNE~\cite{MaatenNov2008} visualization of Gram matrices.
In the figure, we can observe that images of the same lighting condition are grouped together in the style spaces, which validates our assumption.



\begin{figure*}[t]
    \centering
    \vspace{0mm}
    \scalebox{1}{
    \includegraphics[width=\linewidth]{Supplementary/Figures/supple_tsne_gram_v2.pdf} }
\caption{
$t$-SNE visualization of the distributions of Gram matrices computed from low-light and well-lit images.
The Gram matrices are computed from feature maps of the 1st, 2nd, 3rd, and 4th Res Blocks of a CPN model~\cite{chen2018cascaded} pretrained on the ImageNet dataset~\cite{Imagenet}.
In all the visualizations, images of the same lighting condition are clustered together, suggesting that neural styles in the form of Gram matrices encode lighting conditions.
} \label{fig:tsne}
\vspace{0mm}
\end{figure*}

\section{Effect of the Gradient Direction from LUPI Loss}
When training with our LUPI loss, we let the gradient from the loss flow \emph{only} to the student in order to train the student with privileged information from the teacher, i.e., information flows in one direction from the teacher to the student.
In this section, we study the effect of this one-direction strategy of our LUPI loss.
To this end, we prepare three variants of our approach: `T $\rightarrow$ S (Ours)', `T $\leftrightarrow$ S' and `T $\leftarrow$ S'.
`T $\rightarrow$ S (Ours)' is our proposed approach.
`T $\leftrightarrow$ S' allows the gradient from the LUPI loss to flow to both the teacher and student models, i.e., the teacher and student can affect each other.
`T $\leftarrow$ S', on the other hand, allows the gradient to flow only to the teacher.
Table~\ref{tab:lupi_loss} compares the performance of these three variants.
As shown in the table, our approach clearly outperforms the others.
This result implies our one-direction strategy is essential for learning with our LUPI loss.


\begin{table}[t]
\centering
\vspace{0mm}
\caption{
Analysis on the impact of the gradient direction from LUPI loss.}
\vspace{1mm}
\renewcommand{\arraystretch}{1.0}
\scalebox{1}{\begin{tabular}{lcccccc}
\toprule
AP@0.5:0.95     & LL-easy         &  LL-normal       &  LL-hard     &  WL        \\ \midrule
T $\leftrightarrow$ S    &  23.6	    &   14.9	  &  	4.3	 &     	69.3   \\
T $\leftarrow$ S    &  24.0  & 	17.3   & 	6.3	 &  68.5       \\
T $\rightarrow$ S (Ours)   & 29.7   & 18.0  & 6.7       &  69.4      \\ \bottomrule
\end{tabular}}
\vspace{0mm}
\label{tab:lupi_loss}
\end{table}

\section{Effect of Intensity Scaling for Low-light Condition}


As described in the main paper, the average channel intensity of each low-light image is automatically scaled to 0.4 before being fed to the student network.
\Tbl{scaling_effect} shows the performance of Baseline-all and the proposed method trained on original low-light images and scaled low-light images.
In low-light conditions, automatically scaled low-light images significantly improve the performance of both models. However, Baseline-all trained on the scaled low-light images performs much worse in well-lit conditions.

We suspect that, in the case of using original low-light images, the Baseline-all model is biased to the well-lit condition.
It is because well-lit images have large pixel intensities, so the scale of gradient of them is larger than that of original low-light images.
Then, in the case of using scaled low-light images, the Baseline-all model is less biased for the well-lit condition, so the performance on the well-lit condition is decreased.
However, the proposed method is less biased due to the lighting condition invariant features of LSBN and LUPI.
Consequently, Table~\ref{tab:scaling_effect} demonstrates that intensity scaling of low-light images improves the performance of both Baseline-all and our method for low-light conditions.


\begin{table}[t]
\centering
\vspace{0mm}
\caption{
Analysis on impact of scaling for low-light images.}
\vspace{1mm}
\renewcommand{\arraystretch}{1.0}
\scalebox{1}{\begin{tabular}{llcccc}
\toprule
 AP@0.5:0.95  & Method   & LL-easy         &  LL-normal       &  LL-hard     &  WL        \\ \midrule
\multirow{2}{*}{No scaling} & Baseline-all       & 16.3	  & 5.9	  &    1.7	    & 64.2  \\
& Ours       & 26.6  & 	11.9  &  	5.7	  &  68.0       \\ \midrule
\multirow{2}{*}{Scaling}  & Baseline-all       & 24.4	   & 10.7	  & 2.5      &   	30.5      \\
 & Ours       & 29.7   & 18.0  & 6.7       &  69.4      \\  \bottomrule
\end{tabular}}
\vspace{0mm}
\label{tab:scaling_effect}
\end{table}

\section{Additional Analysis on the Style and Feature Gaps Between Different Lighting Conditions}

In the main paper, we compare the Hausdorff distance~\cite{huttenlocher1993comparing} between sets of Gram matrices and features of different lighting conditions.
In this section, we show additional results to investigate the style and feature gaps between pairs of well-lit and low-light images according to the lighting conditions.
To this end, we compute the mean squared error (MSE)~\cite{wang2009mean} distance on features between low-light images and paired well-lit images before and after applying LSBN and LUPI.
Then, we report the average value of MSE distance for total image pairs.
\Fig{feature_distance}(a) presents that the style gaps between low-light and well-lit conditions are reduced by the LUPI loss.
Then, \Fig{feature_distance}(b) also shows that the feature gaps between pairs from different lighting conditions are effectively reduced by both LSBN and LUPI.

\begin{figure*}[t]
    \centering
    \vspace{0mm}
    \scalebox{0.8}{
    \includegraphics[width=\linewidth]{Supplementary/Figures/empirical_analysis_supple_v3.pdf} }
\caption{
Quantitative analysis on the impact of our method.
(a) Style gaps and
(b) feature gaps between low-light (LL) and well-lit (WL) conditions.
The gap between two lighting conditions is measured by the average of the entire image pairs for mean squared error (MSE) distance between image pairs from each condition.
} \label{fig:feature_distance}
\vspace{0mm}
\end{figure*}

\section{Additional Qualitative Results}
\Fig{qual} shows additional qualitative results of Baseline-all, DANN~\cite{dann}+LSBN and our method. This again shows Baseline-all and DANN+LSBN often fail to predict poses, and our method performs better than them.
However, as shown \Fig{failure}, our method cannot predict accurate poses in the low-light-hard conditions.
\Fig{enhance} shows additional examples of enhanced low-light images using LLFlow~\cite{wang2021low}.

\begin{figure*}[t]
    \centering
    \vspace{-2mm}
    \scalebox{1}{
    \includegraphics[width=\linewidth]{Supplementary/Figures/qual_figure_12by5_v1.pdf} }
    \vspace{-7mm}
\caption{
Qualitative results on PID dataset. Each prediction is visualized on the
corresponding low-light image and Ground-truth is visualized on a black image.
(a) Scaled low-light images. (b) Baseline-all. (c) DANN+LSBN. (d) Ours. (e)
Ground-truth.
} \label{fig:qual}
\vspace{-6mm}
\end{figure*}

\begin{figure*}[t]
    \centering
    \vspace{-2mm}
    \scalebox{1}{
    \includegraphics[width=\linewidth]{Supplementary/Figures/failure_figure_4by4_v1.pdf} }
    \vspace{-7mm}
\caption{
Failure cases of our method. (a) Low-light images. (b) Scaled low-light images. (c) Our results. (d) Ground-truth.
} \label{fig:failure}
\vspace{-6mm}
\end{figure*}

\begin{figure*}[t]
    \centering
    \vspace{-2mm}
    \scalebox{1}{
    \includegraphics[width=\linewidth]{Supplementary/Figures/enhance_3by3_v1.pdf} }
    \vspace{-7mm}
\caption{
Qualitative results of LLFlow.
(a) Low-light images. (b) Enhanced
images. (c) Well-lit images.
} \label{fig:enhance}
\vspace{-6mm}
\end{figure*}

\clearpage
%
%
{\small
\bibliographystyle{splncs04}
\bibliography{cvlab_kwak}
}